\documentclass[preprint,12pt,numbers]{elsarticle}
\biboptions{sort&compress}

\usepackage[utf8]{inputenc}
\usepackage[T1]{fontenc}
\usepackage{lmodern}
\usepackage{amsmath,amssymb,amsfonts}
\usepackage{graphicx}
\usepackage{booktabs}
\usepackage{multirow}
\usepackage{array}
\usepackage{makecell}
\usepackage{colortbl}
\usepackage{siunitx}
\usepackage{subcaption}
\usepackage{algorithm}
\usepackage{algorithmic}
\usepackage{setspace}
\usepackage[dvipsnames,table]{xcolor}
\usepackage{xspace}
\usepackage{url}
\usepackage[hidelinks]{hyperref}
\usepackage{cleveref}
\usepackage{placeins}     % \FloatBarrier to stop floats leaking across (sub)sections

\newcommand{\method}{RIM\xspace}
\newcommand{\methodlong}{Retrieval-In-Matching\xspace}

\journal{Knowledge-Based Systems}

\makeatletter
\let\ps@pprintTitle\ps@plain
\makeatother

\begin{document}

\begin{frontmatter}

\title{\method: A Retrieval-In-Matching Framework for
Cross-Domain Global Visual Localization of UAVs\tnoteref{accepted}}
\tnotetext[accepted]{Accepted for publication in \textit{Knowledge-Based Systems}.
Published version:
\href{https://doi.org/10.1016/j.knosys.2026.117027}{doi:10.1016/j.knosys.2026.117027}.
\copyright\ 2026. This manuscript version is made available under the
CC BY-NC-ND 4.0 license
(\url{https://creativecommons.org/licenses/by-nc-nd/4.0/}).}

\author[a,b]{Xin Li\corref{cor1}}
\ead{lixin200@mails.ucas.edu.cn}

\author[a,b]{Siyuan Duan}

\author[a,b]{Shang Wang}

\author[a,b]{Zhimin Mao}

\author[a]{Bingliang Hu}

\author[a]{Geng Zhang\corref{cor1}}
\ead{gzhang@opt.ac.cn}

\cortext[cor1]{Corresponding authors.}

\address[a]{Key Laboratory of Spectral Imaging Technology CAS,
Xi'an Institute of Optics and Precision Mechanics, Chinese Academy
of Sciences, Xi'an, 710119, China}
\address[b]{University of Chinese Academy of Sciences,
Beijing, 100049, China}

\begin{abstract}
% =============================================================
% Abstract — concise KBS version (~230 words).
% The pre-compression version is preserved verbatim in
% sections/00_abstract_original.tex.
% =============================================================
Global visual localization of unmanned aerial vehicles (UAVs) using
remote-sensing reference maps has attracted increasing attention.
However, differences in acquisition time and imaging platform between
UAV and reference imagery introduce substantial cross-domain appearance
and viewpoint shifts, challenging robust
six-degree-of-freedom (6-DoF) pose estimation. To mitigate these shifts,
we render UAV-viewpoint references from Google 3D Tiles across
locations, altitudes, and orientations. A two-stage strategy adapts
SALAD with pose-near positives and geographically distant hard
negatives; local geometric consistency then re-ranks the Top-$K$
candidates. We further propose \methodlong (\method), which freezes the
adapted DINOv2-B retriever and distills a local-descriptor decoder from
its token field and a shallow VGG19 detail stream. One query-side DINOv2-B backbone
forward therefore supports both SALAD retrieval and local description,
eliminating a second foundation-model backbone while preserving the
retrieval descriptors by construction. We evaluate \method zero-shot on
the reconstructed EPFL Urbanscape and self-collected Chang'an Park
datasets, both geographically disjoint from the training data. \method
outperforms ten retrieval baselines. Under the full 3D
distance metric at $25$/$50$\,m, it improves Recall@1 over SALAD by
$8.55$/$13.77$ percentage points on EPFL and $4.45$/$8.94$ points on
Park. At Top-$K{=}5$, the measured online query path through retrieval,
candidate matching, and robust geometric verification takes
$90.8$\,ms: $1.2\times$ faster than the strongest separate
sparse-matching baseline and over $30\times$ faster than RoMa, while
maintaining comparable re-ranking accuracy. These results demonstrate
an efficient UAV global visual localization pipeline under unreliable
satellite navigation.

\end{abstract}


\end{frontmatter}

\singlespacing
\typeout{RIM-ARXIV main: stretch=\baselinestretch; baseline=\the\baselineskip}

% =============================================================
% Section 1 — Introduction
% Verbatim version of the author-provided draft.
% Only changes: (a) \cite{...} placeholders inserted at first
% mention of named methods/datasets, marked with `% TODO`
% comments where the citation key still needs to be confirmed
% in refs.bib;  (b) no prose was rewritten — the §4 revision
% suggestions in the plan are NOT applied here.
% =============================================================
\section{Introduction}
\label{sec:intro}

Reliable navigation of unmanned aerial vehicles (UAVs) relies
predominantly on global navigation satellite systems (GNSS). GNSS
signals are nevertheless vulnerable to multipath and radio-frequency
interference, as well as deliberate jamming or spoofing, which can
severely compromise flight reliability. In GNSS-denied environments,
UAVs require long-term global absolute localization rather than only
low short-term drift. Visual--inertial odometry and simultaneous
localization and mapping (SLAM) have matured substantially as methods
for suppressing short-term drift~\citep{qin2018vinsmono,campos2021orbslam3},
but global relocalization remains more challenging: it must recognize
stable cross-domain visual evidence despite heterogeneous sensors and
long-term appearance or scene changes.

Recent advances in sensing and deep learning have created new
opportunities for UAV global visual localization. Most existing
approaches use widely available two-dimensional (2D) satellite maps,
such as Google or Bing imagery, and divide large remote-sensing images
into patches for retrieval. This formulation commonly simplifies the
UAV's full six-degree-of-freedom (6-DoF) motion by fixing or ignoring
pose components such as flight altitude and pitch. The simplification
is practical for a fixed downward-facing camera and an inherently 2D
reference map, but it discards useful oblique and lateral visual
information. A recent line of work~\citep{zhu2025lodloc} narrows this
gap by aligning images with city-scale Level-of-Detail-1 (LoD-1)
building models. It is, however, restricted to densely built urban
areas and requires a coarse initial pose, limiting its use as a
cold-start relocalization operator when SLAM tracking is lost
(Section~\ref{sec:rel-uav}). Ideally, a vision-only system should
recover the full 6-DoF pose without an initial estimate and operate in
both built and less structured environments.

We use 3D Tiles as a data-construction mechanism for this problem. A
virtual camera samples each target area across locations, altitudes,
and orientations, producing reference views that approximate the UAV
imaging geometry. Reducing avoidable viewpoint discrepancy at the data
level allows learning to focus on the harder shifts caused by sensors,
season, reconstruction, and rendering. The resulting pose-near groups
are structurally compatible with SALAD's place-grouped GSV-Cities
metric-learning interface~\citep{izquierdo2024salad,alibey2022gsvcities},
providing a suitable initialization despite the domain gap between
street-view and aerial imagery.

We formulate retriever adaptation as two complementary stages. Stage~1
transfers similarity-aware representations using pose-near
UAV--reference pairs; Stage~2 performs hard-negative contrastive
refinement. The two stages target distinct error sources. In
particular, after Stage~1 the retriever can still assign high similarity
to images that are visually alike but geographically distant. We
therefore mine these residual negatives once and use them for a short
contrastive refinement.

Because global descriptors alone cannot reliably reject such candidates,
geometric re-ranking of the Top-$K$ results remains essential. Conventional
pipelines, however, run independent global and local encoders for each query,
which adds substantial cost for edge deployment. We therefore propose
\methodlong (\method), an architecture built around a shared DINOv2-B
representation~\citep{oquab2024dinov2}. A frozen SALAD aggregation head
produces the global descriptor from the shared token field, while a distilled
DeDoDe-style decoder~\citep{edstedt2024dedode} combines the shared tokens with
shallow detail features to produce a dense local-descriptor map. Each query consequently requires one shared
DINOv2-B backbone forward rather than two independent foundation-model
encoders. Local decoding and geometric verification remain explicit, while
the adapted DINOv2-B token field is shared by the global aggregation and
local-descriptor branches. This design folds global retrieval into the
query-side feature computation used for matching, eliminating a redundant
foundation-model forward while retaining modular geometric verification.

Three insights from our study directly informed the proposed design:
\begin{itemize}
    \item \textbf{Physical-prior data construction:} viewpoint
    sampling should reduce avoidable geometric discrepancy before
    learning, leaving the model to address the remaining appearance
    shift and perceptual-aliasing errors.
    \item \textbf{Data-efficient foundation-model adaptation:}
    pose-near positives and mined hard negatives address
    complementary cross-domain errors while preserving the general
    representation learned by SALAD.
    \item \textbf{Shared-representation retrieval and matching:}
    global retrieval and local description should reuse one
    foundation-model token field, enabling geometric re-ranking
    without a second independent DINOv2 query encoder.
\end{itemize}

Throughout this paper, \emph{cross-domain} denotes the distribution
shift between \emph{real UAV imagery}, which forms the query domain,
and \emph{rendered UAV-perspective views} generated from a
photogrammetric 3D mesh, which form the reference database. This differs
from the conventional ``UAV$\leftrightarrow$satellite'' pairing, which
we deliberately restructure through UAV-viewpoint sampling.

The main contributions are as follows:
\begin{itemize}
    \item We construct UAV-viewpoint references and pose-near training
    pairs by sampling photogrammetric 3D meshes across UAV positions
    and orientations. This mechanism reduces the cross-view geometric
    gap without assuming exact alignment and supports zero-shot
    evaluation on two geographically disjoint benchmarks.
    \item We propose a two-stage cross-domain adaptation recipe:
    similarity transfer adapts SALAD using pose-near positives, while
    offline hard-negative mining suppresses visually similar but
    geographically distant images that remain after Stage~1.
    \item We propose \methodlong (\method), a shared-representation
    retrieval-and-re-ranking architecture. A frozen SALAD head and a
    distilled local-descriptor decoder consume the same DINOv2-B token
    field, so one shared DINOv2-B backbone forward per query supports
    both global retrieval and local geometric verification.
\end{itemize}

Under the full 3D distance metric, \method improves Recall@1 over SALAD by
$8.55$/$13.77$ percentage points (pp) within $25$/$50$\,m on the reconstructed EPFL
urban benchmark and by $4.45$/$8.94$\,pp on the geographically
disjoint Chang'an Park benchmark. It surpasses ten recent retrieval
baseline families: HLoc/AP-GeM, CLIP, CosPlace, DINOv2~\texttt{[CLS]},
EigenPlaces, AnyLoc, SALAD, SelaVPR++, Game4Loc, and
NetVLAD~(DINOv2). At $K{=}5$, the measured online query path through
retrieval, candidate matching, and robust geometric verification takes
$90.8$\,ms---$1.2\times$ faster than separate DeDoDe and over $30\times$
faster than RoMa, while maintaining comparable
re-ranking accuracy. Section~\ref{sec:exp} reports the complete
benchmarks and ablations.

% =============================================================
% Section 2 — Related Work
% Translated and re-organised from the author-provided Chinese
% draft. Citation keys correspond to refs.bib. The differential
% commentary that the author embedded in the draft (vs. our work)
% has been kept and rephrased as positioning paragraphs at the end
% of the relevant subsections.
% =============================================================
\section{Related Work}
\label{sec:related}

\subsection{UAV Global Visual Localization}
\label{sec:rel-uav}

UAV global visual localization recovers the geographic position of an
aerial query by matching it against a georeferenced database. Early
systems matched GNSS-denied UAV imagery to
satellite tiles~\citep{DBLP:conf/icra/GoforthL19}, learned compact
satellite descriptors~\citep{bianchi2021uav}, or combined cross-view
retrieval with coordinate prediction and geometric matching
~\citep{ahn2021aerialsat,sui2022fast}. University-1652
~\citep{zheng2020university} and UAV-GeoLoc
~\citep{wu2025uavgeoloc} subsequently expanded the scale and viewpoint
diversity of drone geo-localization benchmarks.

Reference-database construction is itself consequential.
\citet{moskalenko2025aerialvprsurvey} showed that map-crop overlap and
zoom level interact with the retrieval method. Recent systems address
viewpoint and data limitations in different ways: UAVVisLoc uses
geometric transformation and coarse-to-fine matching
~\citep{xu2024uavvisloc}; Game4Loc samples arbitrary camera poses in a
virtual world and defines positives by field-of-view overlap
~\citep{ji2025game4loc}; and MMGeo combines image, point-cloud, depth,
and text cues for multimodal retrieval~\citep{ji2025mmgeo}. Game4Loc
provides flexible synthetic data for training and benchmarking, but
its virtual database is not a deployable real-world reference map.
CrossLoc~\citep{epfl_urbanscape} renders pose-calibrated multimodal
observations to supervise a scene-coordinate regressor. In contrast,
our camera-aware rendered views serve both as the searchable reference
database and as adaptation data, reducing geometric mismatch before
learning the remaining appearance shift and geographic confusion
(Section~\ref{sec:method-database}).

Explicit 3D priors provide another route to 6-DoF localization.
LoD-Loc v2~\citep{zhu2025lodloc}, for example, aligns image
silhouettes with city-scale LoD-1 building models and recovers pose
components unavailable from a 2D database. It nevertheless depends on
an urban 3D model and a coarse initial pose. \method instead performs
cold-start global retrieval over camera-aware 3D reference views,
followed by geometric re-ranking and perspective-$n$-point (PnP) pose
estimation, and exposes local correspondences for downstream SLAM or
mapping.

Most relevant to our architecture is Global--Local Visual Localization
(GLVL)~\citep{li2023glvl}, which jointly optimizes retrieval and
matching through a partially shared convolutional neural network
(CNN). GLVL shares the ImageNet-pretrained ResNet50 stem and first
residual stage; the retrieval branch continues through the remaining
ResNet stages and generalized mean (GeM) pooling, while SuperPoint-style
decoders consume the shallow shared feature map. Its retrieval branch
is trained with triplets from the Visual Terrain Relative Navigation
dataset, whereas its matching branch uses synthetic homographies
on remote-sensing images. Alternating between these objectives couples
both tasks through the shared stem.

\method instead decouples optimization while sharing the
complete foundation-model token field at inference. Cross-domain
adaptation first produces the DINOv2-B retriever; its backbone and
SALAD head are then frozen, and only a DeDoDe-style local decoder is
distilled from its token field. Local-feature learning therefore
cannot change the adapted global descriptor. At inference, both
heads reuse the complete DINOv2-B token field rather than an early
CNN stem. GLVL ultimately maps an aerial-image center through a
homography to a 2D coordinate, whereas \method re-ranks Top-$K$ 3D
references with cached local descriptors and robust verification
before estimating a 6-DoF pose with PnP. The contribution is thus
not feature sharing alone, but frozen-retriever preservation,
full-token reuse, and modular local distillation.

\subsection{Visual Place Recognition}
\label{sec:rel-vpr}

A retrieval-based formulation is a natural starting point for UAV
global localization because exhaustive image matching against a
city-scale reference database is computationally prohibitive. Because
UAV and satellite imagery are both acquired by RGB cameras, methods
developed for general visual place recognition (VPR) can transfer,
either directly or through domain adaptation, to aerial localization.

\paragraph{Aggregation-based descriptors}
\citet{arandjelovic2016netvlad} introduced NetVLAD, an end-to-end
trainable extension of vector of locally aggregated descriptors (VLAD)
that has become a de facto VPR baseline.
\citet{noh2017delf} proposed DELF, which combines attention-selected
local descriptors with image-level retrieval while sharing much of the
underlying network. Generalized mean pooling~\citep{radenovic2018gem}
further improves global-descriptor discrimination.
\citet{berton2022cosplace} reformulated large-scale geo-localization as
a classification proxy task in CosPlace, using penultimate-layer
features for retrieval, whereas MixVPR~\citep{alibey2023mixvpr}
cascades feature-mixing layers to capture global relationships across
the spatial elements of feature maps. EigenPlaces
~\citep{berton2023eigenplaces} introduces synthetic viewpoints during
training to improve viewpoint robustness. BoQ
~\citep{alibey2024boq} uses a learnable set of global queries to
aggregate local backbone features through cross-attention.

\paragraph{Foundation-model-based VPR}
Self-supervised vision foundation models have driven a new wave of VPR
research. AnyLoc~\citep{keetha2023anyloc} systematically studies
DINOv2 as a generic feature extractor with several local aggregation
strategies. SALAD~\citep{izquierdo2024salad} combines DINOv2 with
Sinkhorn-based optimal-transport aggregation and achieves strong
performance on standard VPR benchmarks. SelaVPR++
~\citep{lu2025selavprpp} improves retrieval efficiency through binary
coarse search, floating-point re-ranking, and parallel multiscale
convolutional modules. AnyLoc and SALAD highlight a common principle:
strong foundation-model representations benefit from appropriately
curated adaptation data. Stage~1 of our method follows this principle
by constructing paired UAV--remote-sensing groups whose organization is
compatible with GSV-Cities~\citep{alibey2022gsvcities}, the dataset used
to train SALAD. GSV-Cities contains approximately 530K images from 62K
geographically distinct locations and substantial appearance diversity.

\paragraph{Hard-negative mining}
Sample4Geo~\citep{deuser2023sample4geo} introduced a hard-negative
sampling strategy for cross-view geo-localization that uses
geographically nearby locations as positive seeds and mines negatives
by image-embedding similarity. Our fine-tuning recipe follows the same
principle: visually similar but geographically unrelated samples should
be separated in the retrieval space. The key difference is procedural.
Sample4Geo refreshes training embeddings every four epochs through
Dynamic Similarity Sampling, whereas we mine a sparse pool of
residual strict negatives once after Stage~1.

\subsection{Local Feature Matching}
\label{sec:rel-matching}

Global retrieval provides coarse localization, but accurate pose
estimation generally requires geometric refinement, often through PnP
from local feature correspondences. The same matching stage can
re-rank the Top-$K$ candidates and mitigate perceptual aliasing caused
by visually similar but geographically distant scenes. Learned local
features have increasingly surpassed classical methods such as the
scale-invariant feature transform (SIFT), particularly under difficult
cross-domain conditions.

SuperPoint~\citep{detone2018superpoint} jointly predicts pixel-level
interest points and descriptors in one forward pass, preserving the
efficiency of sparse keypoint pipelines while improving accuracy.
SuperGlue~\citep{sarlin2020superglue} enhances SuperPoint matching with
a graph neural network. LoFTR~\citep{sun2021loftr} integrates feature
extraction and correspondence estimation in a detector-free
Transformer and obtains semi-dense matches in low-texture regions.
DKM~\citep{edstedt2023dkm} and RoMa~\citep{edstedt2024roma} instead use
dense correspondence estimation to capture richer pixel-level evidence.

Efficiency-oriented matchers have also narrowed the gap between dense
and sparse matching. ALIKED~\citep{zhao2023aliked} uses a sparse
deformable descriptor head to learn descriptor sampling positions
around detected keypoints. Efficient LoFTR~\citep{wang2024eloftr}
substantially accelerates LoFTR while largely preserving its matching
quality. DeDoDe~\citep{edstedt2024dedode} decouples keypoint detection
from descriptor extraction and trains the two components independently
using multiview co-visibility supervision. MatchAnything
~\citep{he2025matchanything} trains a RoMa-based matcher on large-scale
synthetic multimodal data for cross-modal correspondence estimation.

DeDoDe's modular design motivates its role in \method. Decoupling the
detector from descriptor extraction makes it possible to preserve the
adapted and frozen DINOv2-B retriever, reuse its token field, and train
only a new multiscale descriptor decoder by distillation. The global
retrieval representation obtained in Section~\ref{sec:method-finetune}
is therefore preserved exactly while the same foundation-model tokens
support local matching.

\paragraph{Shared-backbone matchers without global retrieval.}
A closely related line of recent work builds a single backbone
that emits dense per-pixel features suitable for matching,
exemplified by MASt3R and OmniGlue-style dense-foundational
matchers. Although these models also avoid a separate encoder
per candidate, they target \emph{relative} two-view pose
estimation and contain no global-retrieval head. When deployed for
\emph{absolute} global localization, a
separate retrieval step (typically SALAD or NetVLAD) must be
prepended to pick the Top-$K$ reference views: the full system
therefore collapses back into a two-stage pipeline whose
latency scales as $(1 + K)\cdot T_{\text{enc}} + K\cdot
T_{\text{match}}$, identical to the decoupled
``SALAD\,+\,RoMa'' or ``SALAD\,+\,MatchAnything'' baselines we
evaluate in Section~\ref{sec:exp-rerank}. In contrast, \method
integrates retrieval \emph{into} the matching backbone by reusing a
frozen, cross-domain-adapted SALAD head on the same DINOv2-B backbone
and learning only a distilled multiscale descriptor decoder; the global
descriptor and the dense local descriptor map are therefore
produced from one shared DINOv2-B backbone forward, whose cost is paid
once per query independently of $K$. This architectural distinction,
rather than backbone sharing alone, defines retrieval-in-matching.

% =============================================================
% Section 3 — Method (compressed KBS main-manuscript version)
% Detailed derivations, pseudocode, and additional figures are in
% the separately compiled supplementary material.
% =============================================================
\section{Method}
\label{sec:method}

Our method comprises three integrated components. First, a
UAV-viewpoint-sampled database stores each rendered reference view
together with its virtual-camera pose and a pixel-aligned
Earth-Centered, Earth-Fixed (ECEF) coordinate map. Second, a two-stage
adaptation strategy transfers SALAD to real UAV imagery and then
suppresses visually similar but geographically distant references.
Third, \method shares the adapted DINOv2-B token field between a
frozen global aggregation head and a distilled local-descriptor decoder.
At inference, one shared DINOv2-B backbone forward generates the query
representation used by both branches: the global descriptor retrieves
the Top-$K$ references, while the local-descriptor map supports
geometric re-ranking. For the selected reference, geometrically
verified 2D--2D matches index its ECEF coordinate map to form 2D--3D
correspondences. Random sample consensus (RANSAC) with a
perspective-$n$-point (PnP) solver then estimates the absolute 6-DoF
pose. The following subsections formalize the task and describe these
components in the same order.

\subsection{Problem Formulation and Database Definition}
\label{sec:method-formulation}

Given a real UAV query
$I_q\in\mathbb{R}^{H\times W\times3}$, we search a reference
database
\begin{equation}
\mathcal{D}=\{(I_i^{db},\mathbf{T}_i^{db},
\mathbf{M}_i^{ecef})\}_{i=1}^{N},
\end{equation}
where $I_i^{db}$ is a rendered UAV-perspective reference view,
$\mathbf{T}_i^{db}\in SE(3)$ denotes its virtual-camera pose, and
$\mathbf{M}_i^{ecef}\in\mathbb{R}^{H\times W\times3}$ is a
pixel-aligned map storing the ECEF coordinate visible at each reference
pixel. Global descriptors first retrieve the Top-$K$ candidates, which
are re-ranked using local matching and robust geometric verification.
The coordinate map then converts matched reference pixels directly into
3D coordinates for PnP.

\subsection{UAV-Viewpoint Reference Construction}
\label{sec:method-database}

We use 3D Tiles as a data-construction mechanism rather than as
the central algorithmic contribution. A CesiumJS
renderer~\citep{cesiumjs2024} streams Google Photorealistic
3D Tiles~\citep{google3dtiles2024} and exposes a virtual camera
parameterized by ECEF position, yaw, pitch, and roll. Our objective is
not to render a reference image exactly aligned with each query.
Instead, we discretize the expected UAV operating envelope so that the
reference database samples airborne viewpoints more closely than a
fixed nadir satellite map.

For each operating area, we define a horizontal position grid
$\mathcal{G}_{xy}$ and render the Cartesian product
$\mathcal{G}_{xy}\!\times\!\mathcal{H}\!\times\!\Theta\!\times\!\Psi$
of positions, altitudes, pitch angles, and yaw angles; roll is
fixed in the databases used here. The resulting set contains
$N=N_{xy}|\mathcal{H}||\Theta||\Psi|$ views. For an arbitrary query
pose, the nearest sampled camera generally retains nonzero translation
and rotation offsets determined by the spatial and angular sampling
intervals. We therefore describe these references as
\emph{UAV-viewpoint sampled}, not viewpoint aligned. Their purpose is
to reduce geometric viewpoint discrepancy at the data-construction
stage; retrieval and local verification handle the remaining mismatch.
Section~\ref{sec:exp-setup} reports the scene-specific grids.

\begin{figure}[!htbp]
\centering
\includegraphics[width=0.98\linewidth]{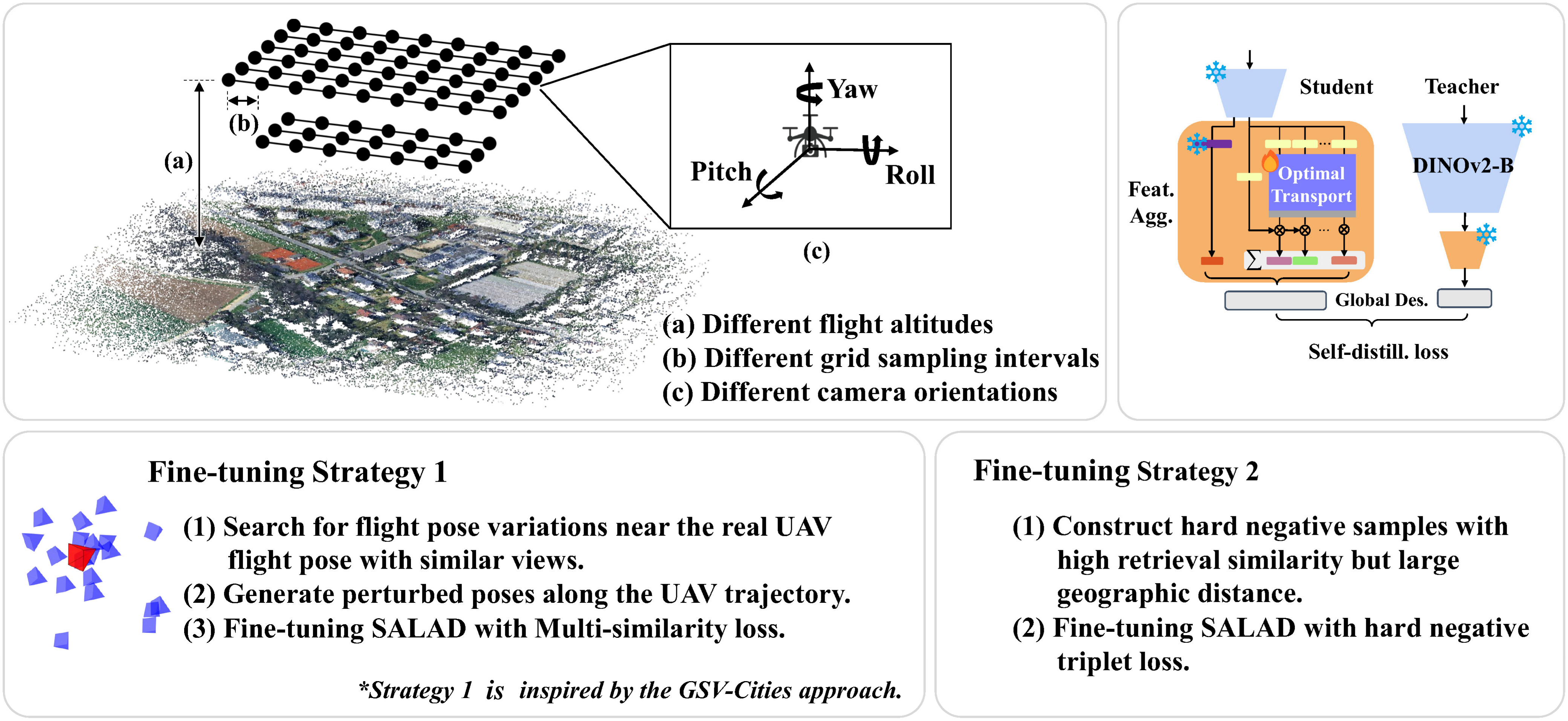}
\caption{UAV-viewpoint reference sampling and two-stage
cross-domain retriever adaptation. Strategy~1 uses pose-near
    positive samples, whereas Strategy~2 suppresses mined hard-negative
samples. The upper-right panel illustrates the Stage~1
    descriptor-level self-distillation regularizer.}
\label{fig:finetune-corpus-samples}
\end{figure}

Figure~\ref{fig:finetune-corpus-samples} connects this sampling
strategy to the subsequent cross-domain retriever adaptation. Its
upper-left panel illustrates the altitude, horizontal-position, and
camera-orientation dimensions.

Each rendered reference stores its RGB image, virtual-camera pose, and
a pixel-aligned ECEF coordinate map obtained by back-projecting the
depth buffer. The retained database stores the full ECEF maps. For
range-based validity filtering, sampled coordinates are expressed
relative to an area-centered origin,
\begin{equation}
\widetilde{\mathbf{M}}_i^{ecef}(u,v)=
\mathbf{M}_i^{ecef}(u,v)-\mathbf{O}_{\mathrm{area}},
\end{equation}
After PnP, the estimated camera center is expressed
relative to the same origin for evaluation.

For training only, we use the known UAV pose to render an auxiliary
set under bounded translation and rotation perturbations. These
query-conditioned renderings are \emph{pose-near positives}, rather
than pose-aligned or pixel-registered pairs, because residual geometric
offsets remain together with sensor, season, mesh-reconstruction, and
rendering differences. The auxiliary views never enter either
evaluation database. By constraining rather than eliminating viewpoint
variation, this construction allows Stage~1 to focus primarily on the
remaining appearance shift.

\subsection{Two-Stage Cross-Domain Retriever Adaptation}
\label{sec:method-finetune}

We adapt SALAD~\citep{izquierdo2024salad} using two sequential stages
designed for limited UAV ground-truth data.
The lower panels of Fig.~\ref{fig:finetune-corpus-samples}
summarize the two stages.

\paragraph{Stage 1: similarity-feature transfer}
Each mini-batch groups a real UAV anchor with its pose-near rendered
positives, and the normalized global descriptors are optimized using
the standard multi-similarity loss~\citep{wang2019multisimilarity}.
Starting from SALAD pretrained on GSV-Cities
~\citep{alibey2022gsvcities}, the complete DINOv2-B backbone and
global-token projection remain frozen; only the cluster, score, and
dust-bin aggregation parameters are updated. A descriptor-level
self-distillation term further limits drift from the original SALAD
descriptor:
\begin{equation}
\mathcal{L}_{\mathrm{stage1}}=
\mathcal{L}_{\mathrm{MS}}+
\lambda_d\lVert v_{\theta}(I)-v_{\theta_0}(I)\rVert_2^2,
\label{eq:salad-stage1}
\end{equation}
where $\theta_0$ denotes the original SALAD weights. This restricted
optimization and descriptor-level regularizer preserve the pretrained
foundation-model representation while adapting SALAD's spatial
assignment and aggregation to the UAV--reference pairs. This global
self-distillation is distinct from the local-descriptor distillation
used by \method below.

\paragraph{Stage 2: visually similar but geographically distant negatives}
Stage~1 still confuses visually similar but spatially unrelated
views. We therefore encode the training gallery once, retrieve each
anchor's Top-$M$ cosine neighbors, and retain negatives that
are both highly similar and farther than a 3D coverage
threshold $\tau_d$. A pose-near rendering provides the positive.
Because this strict filter produces a sparse residual-negative pool, we
oversample the pool and fine-tune for one epoch using a triplet-margin
loss. Unlike online dynamic sampling
\citep{deuser2023sample4geo}, the gallery is encoded only once.
Stage~2 starts from Stage~1 but does not reuse the Stage~1
self-distillation term. Section~\ref{sec:exp-impl} gives the complete
optimization settings.

\subsection{The \method Framework: Retrieval in Matching}
\label{sec:method-rim}

\begin{figure*}[!htbp]
\centering
\includegraphics[width=0.98\linewidth]{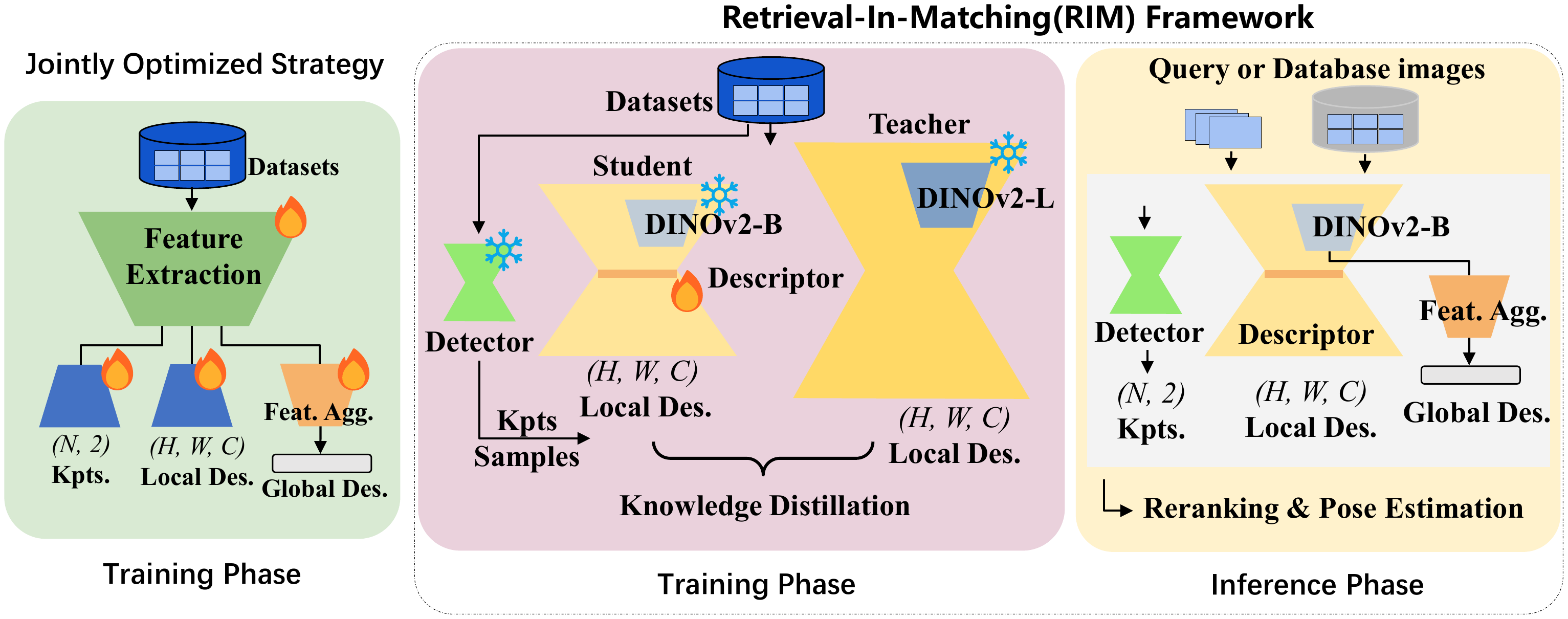}
\caption{Overview of the \methodlong (\method) framework.
\emph{Left:} related jointly optimized global--local architectures.
\emph{Middle:} RIM local-branch adaptation, where a frozen DeDoDe-G
teacher supervises only the student descriptor decoder; all other
components are fixed. \emph{Right:} RIM inference, where one shared
DINOv2-B backbone forward supplies the token field $\mathbf{F}_B$ used
to produce the global descriptor $\mathbf{g}$ and dense local-descriptor
map $\mathbf{D}$; the detector supplies keypoints $\mathbf{K}$.}
\label{fig:rim-framework}
\end{figure*}

Figure~\ref{fig:rim-framework} separates the framework into three
roles. The left panel summarizes the jointly optimized strategy used by
related global--local architectures, in which all components are
adapted together. The middle panel shows RIM adaptation: a frozen
DeDoDe-G teacher supervises only the student descriptor decoder, while
the retrieval representation and remaining operators stay fixed. The
right panel shows the inference graph, where the retrieval-optimized
DINOv2-B representation is computed within the single-image feature
pipeline used for local matching. The global and local outputs
therefore originate from one shared DINOv2-B token field. RIM is not a
sequential composition of SALAD and DeDoDe, but a shared-representation
inference architecture.

\paragraph{Concurrent heads rather than serial modules}
The distinction is clearest in terms of feature ownership. In a serial
SALAD--DeDoDe composition, SALAD uses one DINOv2 token field and the
local model uses another; their representations, weights, and
query-side encoder costs remain independent even when both networks
belong to one software pipeline. In \method, the adapted DINOv2-B
token field is shared system state. The SALAD head aggregates it into a
global descriptor, while the local branch combines the same tokens
with shallow VGG19 features to decode a dense descriptor map. The two
heads are parallel consumers of one representation, not sequential
models. Retrieval therefore does not launch a second DINOv2 encoder;
one token computation branches into global and local outputs before
gallery search and geometric verification.

\paragraph{Shared single-image computation graph}
We make this feature ownership explicit using the notation in
Fig.~\ref{fig:rim-framework}. For an input image $I$, let
$B_{\theta}$ denote the cross-domain-adapted DINOv2-B backbone,
$A_{\phi}$ the global aggregation head, $V_{\omega}$ the shallow
VGG19 feature stream, $\Psi_{\psi}$ the distilled local-descriptor
decoder, and $H_{\eta}$ the DeDoDe L-upright keypoint detector. The
shared and branch-specific computations are
\begin{equation}
\begin{aligned}
\mathbf{F}_{B}(I) &= B_{\theta}(I),
&\mathbf{F}_{V}(I) &= V_{\omega}(I),\\
\mathbf{g}(I) &= A_{\phi}\!\left(\mathbf{F}_{B}(I)\right),
&\mathbf{D}(I) &= \Psi_{\psi}\!\left(
\mathbf{F}_{B}(I),\mathbf{F}_{V}(I)\right),\\
\mathbf{K}(I) &= H_{\eta}(I),
&\mathbf{d}(I) &= \mathcal{S}\!\left(
\mathbf{D}(I),\mathbf{K}(I)\right).
\end{aligned}
\label{eq:rim-shared-graph}
\end{equation}
Here $\mathbf{F}_{B}$ is the token field computed once and consumed by
both heads; $\mathbf{g}$ is the global retrieval descriptor;
$\mathbf{D}$ is the dense local-descriptor map; $\mathbf{K}$ contains
the detected keypoints; and $\mathcal{S}$ samples $\mathbf{D}$ at
$\mathbf{K}$ to produce the local descriptors $\mathbf{d}$. During RIM
distillation, $\theta,\phi,\omega,$ and $\eta$ remain fixed, while only
$\psi$ is optimized by the local-descriptor objective below. Thus,
``shared'' refers specifically to reuse of the DINOv2-B computation;
VGG19, the detector, the decoder, and the aggregation head remain
explicit operators.

This notation also clarifies the ordering implied by the name
Retrieval-In-Matching. Given a query $I_q$ and cached database images
$I_i^{db}$, global retrieval produces a candidate set before mutual
nearest-neighbor (MNN) sparse matching:
\begin{equation}
\begin{aligned}
\mathcal{R}_{K}(I_q)
&=\underset{i}{\operatorname{TopK}}\;
\left\langle\mathbf{g}(I_q),\mathbf{g}(I_i^{db})\right\rangle,\\
\mathcal{M}_{qi}
&=\operatorname{MNN}\!\left(
\mathbf{d}(I_q),\mathbf{d}(I_i^{db})\right),
\qquad i\in\mathcal{R}_{K}(I_q).
\end{aligned}
\label{eq:rim-retrieve-match}
\end{equation}
Retrieval therefore remains prior to candidate matching. The
architectural distinction is that $\mathbf{g}$ is produced within the
same single-image feature-computation pipeline that supplies local
matching, rather than by an independent retrieval backbone. For every
sparse pipeline, database-side tuples
$\{\mathbf{g}(I_i^{db}),\mathbf{K}(I_i^{db}),
\mathbf{d}(I_i^{db})\}$ can be precomputed once and cached offline.

\paragraph{Shared student architecture}
The student retains the cross-domain-adapted DINOv2-B backbone and
SALAD aggregation head, together with a shallow VGG19 stream, the
DeDoDe L-upright detector, and a trainable multiscale descriptor
decoder. The backbone, aggregation head, VGG19 stream, and detector are
frozen; only the approximately $5.6$\,M-parameter decoder is optimized.
The student replaces the DINOv2-L stream of the DeDoDe-G descriptor
teacher~\citep{edstedt2024dedode} with the shared DINOv2-B tokens and
changes the decoder's top input width from $1024$ to $768$. It produces
a $256$-channel dense descriptor map
$\mathcal{D}_S(I)\in\mathbb{R}^{H\times W\times256}$, from which the
unchanged detector selects $N_{\mathrm{kp}}=2000$ locations at
inference. Because the SALAD branch remains frozen, decoder training
cannot alter retrieval accuracy.

\paragraph{Optimization boundary relative to joint global--local training}
The distinction from a partially shared, jointly optimized framework
such as GLVL~\citep{li2023glvl} lies not only in which features are
shared, but also in which task can optimize them. Let $\theta_s$ denote
the shared shallow stem of such a framework. Under alternating
multitask training, both tasks update this stem:
\begin{equation}
\theta_s \leftarrow
\operatorname{Update}(\theta_s,\mathcal{L}_{\mathrm{ret}})
\quad\text{or}\quad
\operatorname{Update}(\theta_s,\mathcal{L}_{\mathrm{match}}).
\label{eq:joint-global-local-update}
\end{equation}
RIM instead keeps the adapted backbone $\theta^\star$ and aggregation
head $\phi^\star$ fixed and trains only the local decoder:
\begin{equation}
\psi^\star=\arg\min_{\psi}\mathcal{L}_{\mathrm{distill}},
\qquad \theta^\star,\phi^\star\ \text{fixed}.
\label{eq:rim-constrained-distillation}
\end{equation}
The VGG19 branch and detector also remain fixed. Consequently, decoder
distillation cannot alter the global descriptor:
\begin{equation}
\mathbf{g}_{\mathrm{after}}(I)
=\mathbf{g}_{\mathrm{before}}(I),\qquad \forall I.
\label{eq:rim-retrieval-invariance}
\end{equation}
This invariant formally distinguishes optimization decoupling from
conventional multitask sharing. It preserves the cross-domain retrieval
gains of the two-stage adaptation while the local decoder learns to
interpret the fixed DINOv2-B token field. At inference, both heads reuse
the complete foundation-model token field $\mathbf{F}_B$, whereas the
shallow VGG19 detail stream remains an explicit local-only operator
(Eq.~\eqref{eq:rim-shared-graph}).

\paragraph{Local-descriptor distillation}
A frozen DeDoDe-G descriptor model, comprising DINOv2-L and VGG19,
provides the teacher map $\mathcal{D}_T$. On channel-normalized maps,
we combine point-wise mean-squared error (MSE), cosine dissimilarity,
and sign consistency, with $w(u,v)$ up-weighting the hardest spatial
locations:
\begin{equation}
\begin{aligned}
\mathcal{L}_{\mathrm{distill}}
=\frac{1}{HW}\sum_{u,v}w(u,v)\big[
&\lambda_{\mathrm{mse}}\|\widetilde{\mathcal{D}}_S-
 \widetilde{\mathcal{D}}_T\|_2^2\\
&+\lambda_{\cos}(1-\cos(\widetilde{\mathcal{D}}_S,
 \widetilde{\mathcal{D}}_T))\\
&+\lambda_{\mathrm{sgn}}\operatorname{ReLU}(
 -\widetilde{\mathcal{D}}_S\odot
  \widetilde{\mathcal{D}}_T)\big]_{u,v}.
\end{aligned}
\label{eq:rim-distill}
\end{equation}
This teacher--student objective acts only on the local-descriptor maps;
because the retriever remains frozen, local distillation cannot alter
the adapted global descriptor.

The three terms play complementary roles. MSE transfers the teacher's
descriptor geometry after channel normalization, cosine loss preserves
angular neighborhoods used by MNN search, and sign consistency
penalizes dimension-wise inversions that can leave average cosine
similarity deceptively stable. Hard spatial weighting concentrates the
decoder's limited capacity on locations where the smaller DINOv2-B
student disagrees most with the DINOv2-L teacher. The teacher is used
only during training; inference retains the student decoder, frozen
global head, shared DINOv2-B backbone, and shallow detail stream.

\paragraph{Single-pass inference and computational cost}
For each query, the shared DINOv2-B backbone is evaluated once; the
explicit heads then produce the SALAD global descriptor, DeDoDe
keypoints, and distilled dense descriptors. Reference descriptors are
cached offline, and MNN matching followed by robust geometric
verification provides the re-ranking score. Let $t_g$, $t_l$, $t_r$,
and $t_m$ denote the costs of separate global encoding, local encoding,
gallery retrieval, and per-candidate matching, respectively. A
decoupled sparse pipeline requires
\begin{equation}
T_{\mathrm{sep}}(K)=t_g+t_r+t_l+K\,t_m,
\label{eq:rim-cost-baselines}
\end{equation}
whereas \method requires
\begin{equation}
T_{\mathrm{RIM}}(K)=t_{\mathrm{shared}}+t_{\mathrm{head}}
+t_r+K\,t_{\mathrm{MNN}}.
\label{eq:rim-cost-ours}
\end{equation}
The saving is primarily $K$-independent: the second query-side DINOv2
encoding is eliminated, and DINOv2-L is replaced by the smaller shared
DINOv2-B representation. For dense matchers, \method also avoids a
full pairwise network forward for each candidate. The cost equations
predict two behaviors. First, \method and separate sparse DeDoDe scale
similarly with $K$ because both use cached reference descriptors and
inexpensive MNN matching; their difference is mainly the constant
query-side encoding cost. Second, dense pairwise matchers scale much
more steeply because every additional candidate invokes another
matcher-network forward. RIM reduces the former constant cost while
preserving the favorable scaling of the sparse pipeline, so its
efficiency gain persists across re-ranking depths rather than depending
on a particular $K$.

\begin{figure*}[!htbp]
\centering
\includegraphics[width=0.92\linewidth]{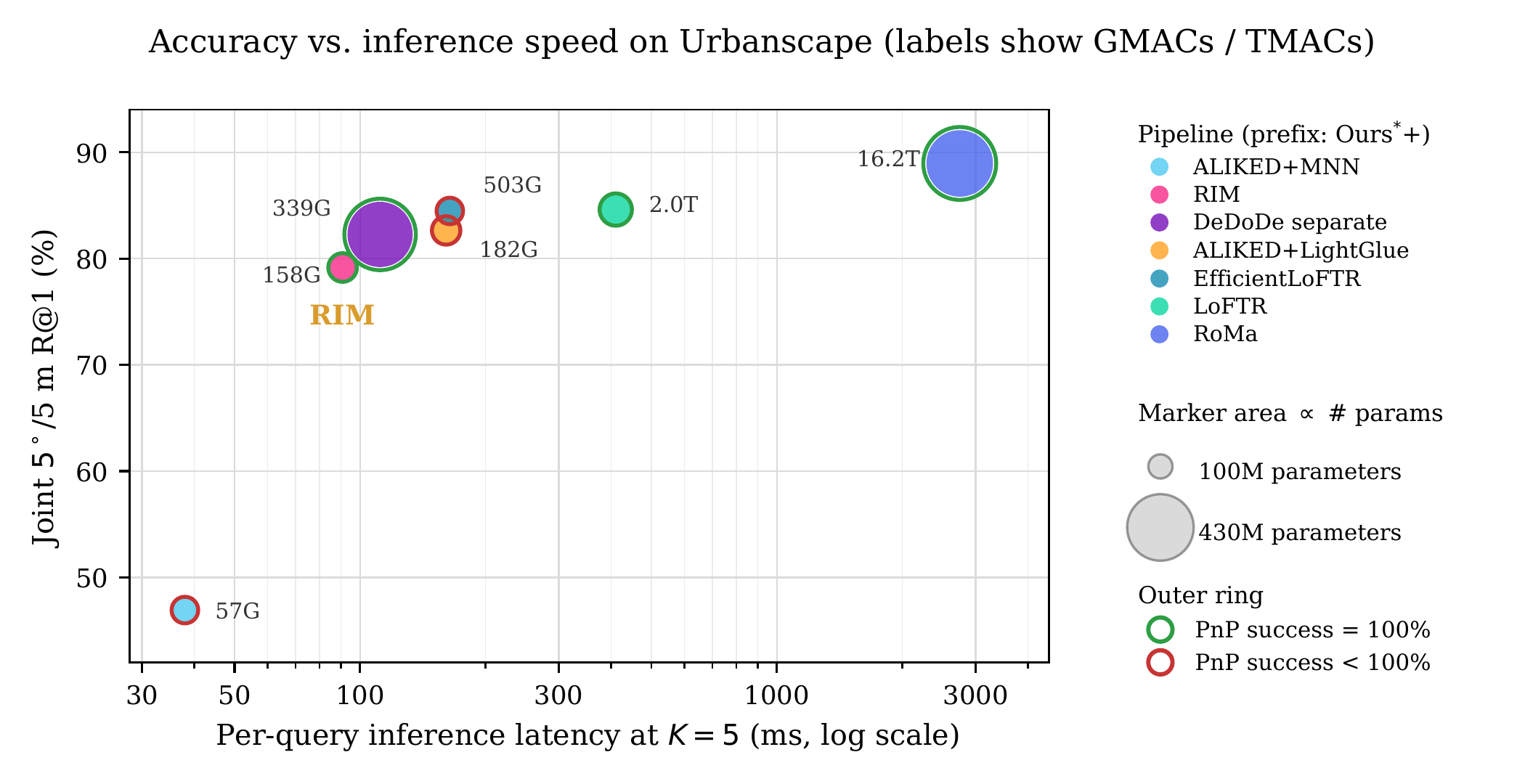}
\caption{Accuracy--speed--size Pareto view on Urbanscape. Each
marker is one selected Ours$^{\ast}$\,+\,matcher pipeline with
audited parameter and multiply--accumulate counts. Horizontal:
per-query latency at $K{=}5$ from Table~\ref{tab:latency}
(32-bit floating point (FP32); RTX 3090; log scale). Vertical: joint
$5^{\circ}\!/5$\,m Recall@1 (R@1) from Table~\ref{tab:pnp-urbanscape}.
Marker area $\propto$ query-side parameter footprint. The adjacent label
gives the $K{=}5$ compute budget in billions or trillions of
multiply--accumulate operations (GMACs/TMACs) at $322{\times}322$ and
$N{=}2000$ keypoints. Green ring:
perspective-$n$-point (PnP) solver return $=100\%$; red ring: $<\!100\%$. \method
($\sim\!119$ million parameters, $\sim\!158$\,GMACs, $90.8$\,ms) sits
at the upper-left of the plot: it is the fastest of the four pipelines
with a $100\%$ solver-return rate and uses less than half the compute of
the next such competitor. Its joint pose score is within
approximately $3$ percentage points of Ours$^{\ast}$\,+\,DeDoDe (separate), the
original distillation teacher ($\sim\!339$\,GMACs). Dense matchers
(LoFTR $\sim\!2$\,TMACs, RoMa $\sim\!16$\,TMACs) sit one to
two orders of magnitude further to the right.}
\label{fig:accuracy-speed-compute}
\end{figure*}

\paragraph{Deployment implications}
Representation sharing also reduces runtime model residency. A
separate SALAD--DeDoDe system retains both the retrieval-oriented
DINOv2-B and matching-oriented DINOv2-L backbones, together with their
heads. The deployed \method model removes the DeDoDe teacher: the
shared DINOv2-B backbone is stored once, alongside the shallow detail
stream and $5.6$\,M-parameter student decoder. Query processing thus
does not retain activations for two independent foundation-model
graphs. This matters onboard because model residency and peak
activation pressure compete with mapping, planning, and control for
the same accelerator resources. RIM's systems advantage should
therefore be assessed jointly in latency, computation, and parameter
footprint (Fig.~\ref{fig:accuracy-speed-compute}).

\paragraph{Scope of the single-pass claim}
``Single pass'' in this paper refers specifically to one \emph{shared
DINOv2-B backbone forward per query}. The shallow VGG19 stream,
descriptor decoder, detector, gallery search, MNN comparison,
geometric verification, and PnP remain explicit operations.
Reference-side global and local descriptors are precomputed
offline for all sparse pipelines, including the baselines, and
are not counted as an online saving unique to \method. This
definition makes the comparison conservative: the reported gain comes
from eliminating duplicated query-side representation and using the
smaller distilled backbone, not from excluding database work or
geometric reasoning.
Section~\ref{sec:exp-latency} tests this prediction.

\subsection{From Matching to Absolute Pose}
\label{sec:method-pose}

After re-ranking, inlier matches
$\{(\mathbf{u}_q^{(k)},\mathbf{u}_{db}^{(k)})\}$ are converted
into 2D--3D correspondences by indexing the reference map,
$\mathbf{X}_{ecef}^{(k)}=
\widetilde{\mathbf{M}}_*^{ecef}(\mathbf{u}_{db}^{(k)})+
\mathbf{O}_{\mathrm{area}}$.
RANSAC-PnP then estimates
\begin{equation}
\mathbf{R}_q,\mathbf{t}_q
=\arg\min_{\mathbf{R},\mathbf{t}}\sum_k
\rho\!\left(\|\mathbf{u}_q^{(k)}-
\pi(\mathbf{R}\mathbf{X}_{ecef}^{(k)}+\mathbf{t})
\|_{\Sigma}^{2}\right).
\label{eq:pnp}
\end{equation}
The estimated camera center is then translated into the area-centered
evaluation frame. This standard PnP stage is not claimed as a separate contribution; the
pixel-aligned coordinate map enables it without depth estimation,
triangulation, or an initial pose.

% =============================================================
% Section 4 — Experiments (compressed KBS main-manuscript version)
% Extended ablations, capture details, and qualitative results are
% provided in the separately compiled supplementary material.
% =============================================================
\section{Experiments}
\label{sec:exp}

\subsection{Datasets and Evaluation Protocol}
\label{sec:exp-setup}

We evaluate zero-shot on two test geographies disjoint from the
Urbanscape Out-of-Place (Urbanscape-OOP) fine-tuning set.
\emph{Urbanscape} is the primary benchmark: $3{,}158$ real EPFL UAV
images with 6-DoF ground truth (GT) are queried against $17{,}400$
UAV-perspective views rendered from Google 3D Tiles. Its urban
geometry and broad variation in altitude and attitude introduce
substantial viewpoint and appearance shifts. \emph{Chang'an Park} is
a self-collected secondary benchmark dominated by vegetation, water,
and seasonal change. Its references are rendered from an independently
reconstructed structure-from-motion (SfM) mesh built from imagery
acquired approximately one year earlier. We exclude only $68$
construction frames depicting content absent from the reference,
leaving $2{,}315$ of $2{,}383$ queries. Query translation is obtained
from dual-frequency real-time kinematic positioning, while
orientation is taken from exchangeable image file format
metadata. A single constant ECEF translation aligns the vertical datums
of the query and mesh coordinate systems; rotations remain unchanged.
Supplementary~\ref{app:park-gt} provides the ground-truth audit.

\begin{figure*}[!htbp]
\centering
\includegraphics[width=0.98\linewidth]{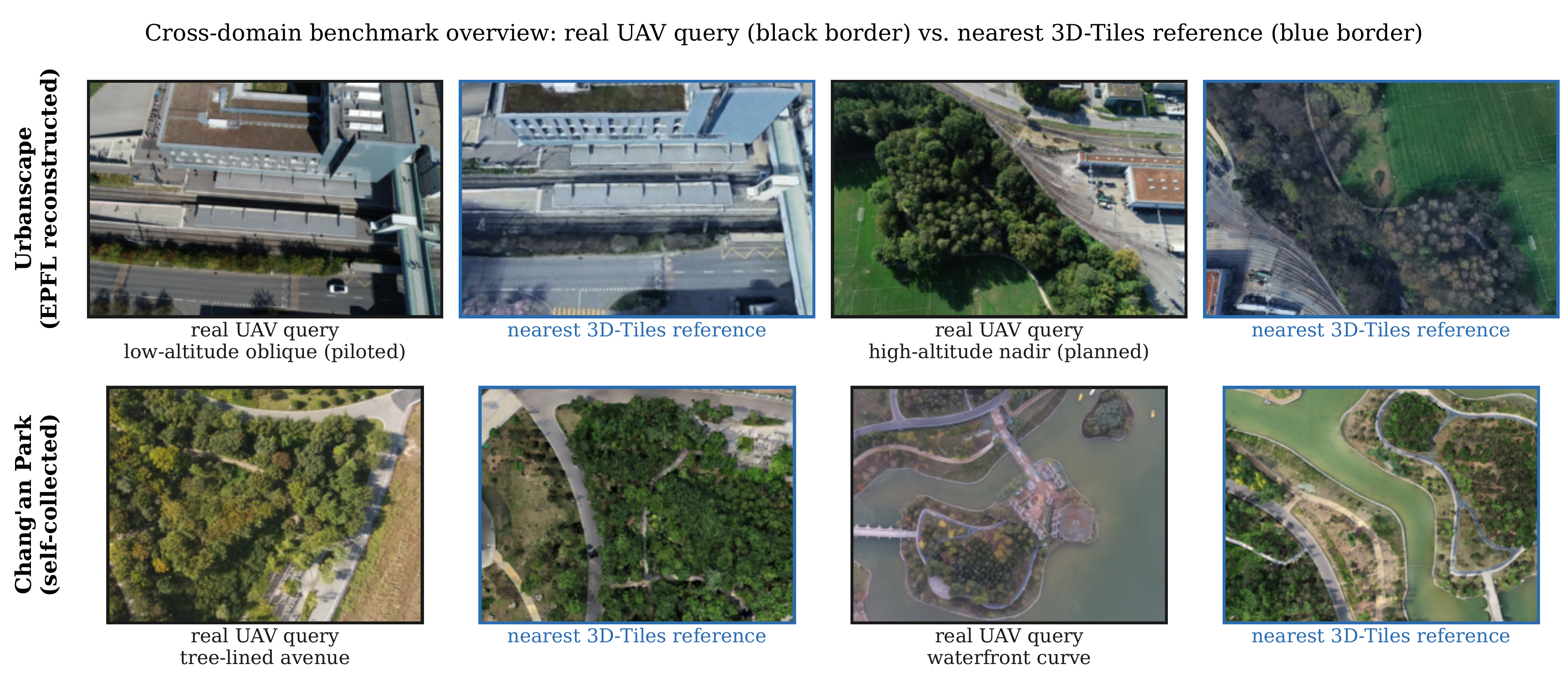}
\caption{Cross-domain benchmark overview: representative real UAV
queries (black borders) next to their nearest
UAV-perspective reference tile (blue border) produced by the
grid sampler of Section~\ref{sec:method-database}.
\emph{Top, Urbanscape (EPFL):} a piloted oblique flight
\emph{vs.}\ a nadir tile from a planned high-altitude grid,
illustrating the substantial viewpoint variation that a single
photogrammetric mesh must accommodate.
\emph{Bottom, Chang'an Park:} a tree-lined avenue and a
waterfront curve, illustrating the vegetation\,+\,water
low-landmark regime where the reference and query imagery were acquired
at least one year apart. The two benchmarks span
qualitatively different axes of cross-domain difficulty
(building clutter \emph{vs.}\ vegetation, viewpoint shift
\emph{vs.}\ temporal drift).}
\label{fig:dataset-overview}
\end{figure*}

Both evaluation databases use a nominal $20$\,m horizontal grid with
zero roll. Urbanscape samples altitudes
$\{500,575,650,725,800\}$\,m above sea level at a fixed
$-75^\circ$ pitch and yaws
$\{0^\circ,90^\circ,180^\circ,270^\circ\}$. Chang'an Park
samples altitudes $\{540,565,590\}$\,m, pitches
$\{-90^\circ,-60^\circ\}$, and yaws
$\{0^\circ,90^\circ,180^\circ,270^\circ\}$.
These discrete grids provide viewpoint coverage but do not assume that
a query and reference share identical extrinsics.
Table~\ref{tab:dataset-stats} summarizes the dataset scales,
while Fig.~\ref{fig:dataset-overview} illustrates the two test
domains and their query--reference shifts.

\begin{table}[!htbp]
\centering
\caption{Scale of the three datasets used in this paper. The
fine-tuning split (Urbanscape-OOP) is geographically
\emph{disjoint} from both test scenes. \emph{Real-UAV queries}
denotes the number of onboard red--green--blue (RGB) frames with
six-degree-of-freedom (6-DoF) ground-truth (GT) poses;
\emph{Reference database (DB) tiles} denotes the number of UAV-perspective
renderings produced by the grid sampler of
Section~\ref{sec:method-database} on top of Google
Photorealistic 3D Tiles (Urbanscape, Urbanscape-OOP) or a
self-reconstructed structure-from-motion (SfM) mesh (Chang'an Park).
\emph{Scene area} is the convex hull of the reference grid in
the local east--north--up plane.
\emph{Flight altitude} is the range of World Geodetic System 1984
ellipsoidal heights of the real-UAV poses.}
\label{tab:dataset-stats}
\setlength{\tabcolsep}{4pt}
\resizebox{\columnwidth}{!}{%
\begin{tabular}{lcrrrcr}
\toprule
\textbf{Dataset} & \textbf{Role} & \textbf{Real-UAV} & \textbf{Reference} & \textbf{Scene} & \textbf{Flight altitude} & \textbf{Trajectory} \\
                &               & \textbf{queries}  & \textbf{DB tiles}  & \textbf{area (km$^2$)} & \textbf{(m, range)} & \textbf{(km)} \\
\midrule
Urbanscape-OOP        & fine-tuning      & 1{,}360 & 49{,}920 & 0.63 & 0--204 & 14.3 \\
Urbanscape (EPFL)     & test (zero-shot) & 3{,}158 & 17{,}400 & 0.34 & 0--83  & 39.9 \\
Chang'an Park (ours)  & test (zero-shot) & 2{,}383 & 26{,}568 & 2.52 & 0--131 & 20.9 \\
\bottomrule
\end{tabular}}
\end{table}

The full on-disk reference-database footprint is 80.6\,GiB for
Urbanscape (13.4\,GiB RGB plus 67.2\,GiB pixel-aligned coordinate
maps) and 107.8\,GiB for Park (16.6\,GiB plus 91.2\,GiB,
respectively).
These are persistent-storage footprints, not resident RAM or GPU-memory
requirements: final PnP accesses only the selected candidate's coordinate
map on demand.

Retrieval is evaluated using Recall@$K$ ($K\in\{1,5\}$) under
3D Euclidean thresholds of $25$ and $50$\,m. Planar XY results
are reported in Supplementary~\ref{app:xy-tables}.
Pose evaluation reports the PnP solver-return rate, conditional median
translation and rotation errors over finite solver returns, and joint
$5$\,m/$5^\circ$ Recall over all queries.

The experimental platform
comprised an NVIDIA RTX~3090 GPU (24\,GB), dual Intel Xeon Platinum 8260
CPUs (2.40\,GHz; 48 physical cores in total), and 512\,GB RAM.
Retrieval tables precede re-ranking so that retriever and matcher
contributions remain separable.

\subsection{Implementation Details}
\label{sec:exp-impl}

Stage~1 uses DINOv2-B SALAD at $322{\times}322$, optimized with AdamW
($6{\times}10^{-5}$), a batch size of $32$, and one epoch. The complete
DINOv2-B backbone and global-token projection are frozen; only the
SALAD cluster, score, and dust-bin aggregation parameters are
updated, with $\lambda_d=0.5$ for the descriptor-level
self-distillation term. Stage~2 continues for
one epoch on the mined triplets using a margin of $0.3$, AdamW
($7{\times}10^{-5}$), a batch size of $16$, and $10{\times}$
oversampling. \method distillation uses $518{\times}518$
inputs, a batch size of $4$, AdamW ($10^{-4}$), and
$(\lambda_{\mathrm{mse}},\lambda_{\cos},\lambda_{\mathrm{sgn}})
=(1.0,0.5,0.1)$ with a $20\%$ hard-pixel fraction. Only the
$5.6$\,M-parameter decoder is optimized; the best validation
checkpoint from at most $100$ epochs is retained.

\subsection{Cross-Domain Retrieval}
\label{sec:exp-main}

We compare ten baseline method families spanning aggregation,
foundation-model, transformer, and UAV-specific retrieval:
HLoc/AP-GeM, CLIP, CosPlace, DINOv2~\texttt{[CLS]}, EigenPlaces,
AnyLoc, SALAD, SelaVPR++, Game4Loc, and NetVLAD~(DINOv2).
Nine rows use released checkpoints; to separate data access from
architecture, NetVLAD~(DINOv2) and an additional SelaVPR++
variant are trained on the same pairs and two stages as our retriever.
Tables~\ref{tab:rec-urbanscape-3d} and
\ref{tab:rec-park-3d} use an asterisk for these trained
variants; \textbf{Ours}$^\ast$ is the adapted SALAD retriever
before geometric re-ranking.
All Ours$^\ast$ retrieval-only, geometric re-ranking, PnP, and
controlled-ablation results use the same two-stage adapted
retriever and the Top-$K$ candidate lists it produces.

\begin{table*}[!htbp]
\centering
\caption{Recall@$K$ (\%) on the EPFL urban (\textit{Urbanscape})
benchmark with the full 3D Euclidean distance metric at
tolerance radii $\tau\in\{25,50\}$\,m. Best per column in
\textbf{bold}, runner-up \underline{underlined}. ``$\ast$''
marks retrievers whose head has been retrained on our pose-near
UAV--reference pairs
(Section~\ref{sec:method-database}); Game4Loc is evaluated
\emph{zero-shot} from the authors' GTA-UAV weights, while the
unstarred general visual-place-recognition rows use officially released weights.
The planar (XY) counterpart is in
Supplementary~\ref{app:xy-tables}.}
\label{tab:rec-urbanscape-3d}
\setlength{\tabcolsep}{4pt}
\scriptsize
\resizebox{\linewidth}{!}{%
\begin{tabular}{l|c|cc|cc}
\toprule
& & \multicolumn{2}{c|}{$\tau = 25$\,m} & \multicolumn{2}{c}{$\tau = 50$\,m} \\
Method & Dimension & R@1 & R@5 & R@1 & R@5 \\
\midrule
HLoc (AP-GeM)~\citep{sarlin2019hloc,revaud2019apgem} & $2048$  & 16.75 & 33.06 & 51.39 & 73.37 \\
CLIP~\citep{radford2021clip}                          & $768$   & 2.53  & 6.65  & 21.34 & 27.07 \\
CosPlace~\citep{berton2022cosplace}                   & $2048$  & 14.38 & 25.68 & 38.22 & 60.73 \\
DINOv2 \texttt{[CLS]}~\citep{oquab2024dinov2}         & $768$   & 8.39  & 17.80 & 29.48 & 51.71 \\
EigenPlaces~\citep{berton2023eigenplaces}             & $2048$  & 21.31 & 34.83 & 55.26 & 76.22 \\
AnyLoc~\citep{keetha2023anyloc}                       & $49152$ & 20.49 & 35.50 & 50.66 & 77.07 \\
SALAD~\citep{izquierdo2024salad}                      & $8448$  & 18.43 & 31.44 & 53.93 & 76.06 \\
SelaVPR++~\citep{lu2025selavprpp}                     & $2048$  & 18.52 & 34.67 & 54.84 & 78.97 \\
Game4Loc (zero-shot)~\citep{ji2025game4loc}               & $768$   & 15.90 & 31.13 & 39.58 & 62.95 \\
\midrule
NetVLAD (DINOv2)$^{\ast}$~\citep{arandjelovic2016netvlad} & $49152$ & 17.48 & 29.39 & 42.88 & 64.63 \\
SelaVPR++$^{\ast}$~\citep{lu2025selavprpp}                & $2048$  & \underline{22.17} & \underline{38.32} & \underline{58.49} & \underline{80.78} \\
\rowcolor{gray!12}
\textbf{Ours}$^{\ast}$                                    & $8448$
  & \textbf{26.98}\,\scriptsize{\textcolor{red}{$+8.55$}}
  & \textbf{42.31}\,\scriptsize{\textcolor{red}{$+10.87$}}
  & \textbf{67.70}\,\scriptsize{\textcolor{red}{$+13.77$}}
  & \textbf{87.40}\,\scriptsize{\textcolor{red}{$+11.34$}} \\
\bottomrule
\end{tabular}}\\[2pt]
\parbox{0.95\linewidth}{\centering\scriptsize Red deltas: gain of
\textbf{Ours}$^{\ast}$ over vanilla SALAD. ``$\ast$'' marks
models whose retrieval head is retrained on our pose-near
UAV--reference pairs under the same
Stage\,1\,$\to$\,Stage\,2 recipe as Ours$^{\ast}$.}
\end{table*}

\begin{table*}[!htbp]
\centering
\caption{Recall@$K$ (\%) on the Chang'an Park benchmark, 3D
Euclidean distance metric, on the $2{,}315$-query
construction-filtered set (matching
Table~\ref{tab:park-attribution}). Conventions follow
Table~\ref{tab:rec-urbanscape-3d}; the planar (XY) counterpart
is in Supplementary~\ref{app:xy-tables}.}
\label{tab:rec-park-3d}
\setlength{\tabcolsep}{4pt}
\scriptsize
\resizebox{\linewidth}{!}{%
\begin{tabular}{l|c|cc|cc}
\toprule
& & \multicolumn{2}{c|}{$\tau = 25$\,m} & \multicolumn{2}{c}{$\tau = 50$\,m} \\
Method & Dimension & R@1 & R@5 & R@1 & R@5 \\
\midrule
HLoc (AP-GeM)~\citep{sarlin2019hloc,revaud2019apgem} & $2048$  & 10.67 & 21.81 & 22.07 & 41.99 \\
CLIP~\citep{radford2021clip}                          & $768$   & 2.25  & 7.56  & 7.95  & 26.39 \\
CosPlace~\citep{berton2022cosplace}                   & $2048$  & 5.87  & 13.91 & 13.09 & 29.94 \\
DINOv2 \texttt{[CLS]}~\citep{oquab2024dinov2}         & $768$   & 6.91  & 14.99 & 20.43 & 39.57 \\
EigenPlaces~\citep{berton2023eigenplaces}             & $2048$  & 9.72  & 20.09 & 19.96 & 39.09 \\
AnyLoc~\citep{keetha2023anyloc}                       & $49152$ & 10.54 & \underline{25.83} & 23.07 & 51.19 \\
SALAD~\citep{izquierdo2024salad}                      & $8448$  & 10.63 & 24.45 & 23.24 & 51.53 \\
SelaVPR++~\citep{lu2025selavprpp}                     & $2048$  & \underline{12.10} & 24.71 & \underline{30.89} & \underline{53.78} \\
Game4Loc (zero-shot)~\citep{ji2025game4loc}               & $768$   & 10.41 & 21.77 & 20.04 & 42.59 \\
\midrule
NetVLAD (DINOv2)$^{\ast}$~\citep{arandjelovic2016netvlad} & $49152$ & 7.99  & 17.06 & 22.51 & 40.13 \\
SelaVPR++$^{\ast}$~\citep{lu2025selavprpp}                & $2048$  & 9.24  & 21.43 & 27.78 & 49.72 \\
\rowcolor{gray!12}
\textbf{Ours}$^{\ast}$                                    & $8448$
  & \textbf{15.08}\,\scriptsize{\textcolor{red}{$+4.45$}}
  & \textbf{31.36}\,\scriptsize{\textcolor{red}{$+6.91$}}
  & \textbf{32.18}\,\scriptsize{\textcolor{red}{$+8.94$}}
  & \textbf{60.00}\,\scriptsize{\textcolor{red}{$+8.47$}} \\
\bottomrule
\end{tabular}}\\[2pt]
\parbox{0.95\linewidth}{\centering\scriptsize Red deltas: gain of
\textbf{Ours}$^{\ast}$ over vanilla SALAD.}
\end{table*}

Ours$^\ast$ ranks first in every 3D column on both unseen
regions. On Urbanscape it improves vanilla SALAD by
$8.55$\,pp at R@1$_{25}$ and $13.77$\,pp at R@1$_{50}$; on
Park the gains are $4.45$ and $8.94$\,pp. Training other heads
on the same pairs does not reproduce these gains, indicating
that they arise from the interaction between SALAD's
aggregation and the two-stage recipe rather than generic
exposure to the rendered data. The complete XY tables,
cross-backbone ablation, aggregate plots, and qualitative
examples are in Supplementary
\ref{app:xy-tables}, \ref{app:cross-backbone}, and
\ref{app:additional-figures}.

\subsection{Geometric Re-ranking and Pose}
\label{sec:exp-rerank-importance}
\label{sec:exp-rerank}

Geometric evidence is necessary because global descriptors
retain perceptual aliasing. Table~\ref{tab:rerank-importance}
isolates this effect before comparing matchers.

\begin{table}[!htbp]
\centering
\caption{Quantitative effect of geometric re-ranking on the
Urbanscape test queries for planar (XY) and full 3D distance, under the same
Ours$^{\ast}$\,+\,\method pipeline. The
retrieval ranking is produced by the global descriptor; the
re-ranked order is produced by the 2D--2D fundamental-matrix
inlier count obtained with Universal Sample Consensus and
Marginalizing Sample Consensus scoring (USAC-MAGSAC) on the local descriptor
matches (Section~\ref{sec:method-rim}). The perspective-$n$-point (PnP) inlier row is a
downstream diagnostic measured on the selected Top-1.}
\label{tab:rerank-importance}
\setlength{\tabcolsep}{4pt}
\scriptsize
\resizebox{\linewidth}{!}{%
\begin{tabular}{l|cc|cc}
\toprule
& \multicolumn{2}{c|}{XY distance (\%)} & \multicolumn{2}{c}{3D distance (\%)} \\
Metric & $\tau = 25$\,m & $\tau = 50$\,m & $\tau = 25$\,m & $\tau = 50$\,m \\
\midrule
R@1 (retrieval only)     & 44.87 & 85.31 & 26.98 & 67.70 \\
R@1 (after re-ranking)   & \textbf{52.41} & \textbf{88.38} & \textbf{34.26} & \textbf{70.80} \\
$\Delta$                 & \textcolor{OliveGreen}{$+7.54$}
                         & \textcolor{OliveGreen}{$+3.07$}
                         & \textcolor{OliveGreen}{$+7.28$}
                         & \textcolor{OliveGreen}{$+3.10$} \\
\midrule
Median Top-1 XY error (m) before / after & \multicolumn{2}{c|}{$27.69 \rightarrow 23.65$}
                                         & \multicolumn{2}{c}{median 3D: $43.79 \rightarrow 43.20$\,m} \\
Mean PnP inliers before / after          & \multicolumn{4}{c}{$210 \rightarrow 295$ on the Ours$^{\ast}$\,+\,\method pipeline} \\
\bottomrule
\end{tabular}}
\end{table}

At $25$\,m, re-ranking raises R@1 by $7.54$\,pp in XY and
$7.28$\,pp in 3D while increasing the mean PnP inlier count
from $210$ to $295$. The corresponding cases are visualized in
Supplementary Fig.~\ref{fig:rerank-qual}.

All pipelines in Tables~\ref{tab:rerank-urbanscape} and
\ref{tab:pnp-urbanscape} use the same Ours$^\ast$ candidates
and PnP solver; only the matching front-end changes.

\begin{table*}[!htbp]
\centering
\caption{Top-$K{=}5$ geometric re-ranking on EPFL urban (Recall@1, R@1,
\%). Row~1 is Ours$^{\ast}$ retrieval only;
remaining rows
apply the indicated matcher on the same Top-5 candidate list.
R@5 is identical to the retrieval-only row ($74.16$ / $94.87$
at $\tau{=}25/50$\,m XY, $42.31$ / $87.40$ at $\tau{=}25/50$\,m
3D) since re-ranking only permutes the retrieved set. All
matchers run at $322{\times}322$ with Universal Sample Consensus and
Marginalizing Sample Consensus scoring (USAC-MAGSAC). Best and
second-best results are bold and underlined, respectively; tied
best values are all bold.}
\label{tab:rerank-urbanscape}
\setlength{\tabcolsep}{4pt}
\scriptsize
\resizebox{\linewidth}{!}{%
\begin{tabular}{l|cc||cc}
\toprule
& \multicolumn{2}{c||}{XY distance, R@1 (\%)}
& \multicolumn{2}{c}{3D distance, R@1 (\%)} \\
Pipeline (retriever $+$ matcher)
   & $\tau=25$\,m & $\tau=50$\,m
   & $\tau=25$\,m & $\tau=50$\,m \\
\midrule
Ours$^{\ast}$ (retrieval only, no rerank)     & 44.87 & 85.31 & 26.98 & 67.70 \\
\midrule
Ours$^{\ast}$\,+\,DeDoDe (separate)~\citep{edstedt2024dedode}
                                                & \underline{53.36} & 88.47 & 34.45 & 71.47 \\
Ours$^{\ast}$\,+\,ALIKED\,+\,LightGlue~\citep{zhao2023aliked,lindenberger2023lightglue}
                                                & 51.52 & 89.14 & 32.43 & 69.79 \\
Ours$^{\ast}$\,+\,ALIKED\,+\,MNN~\citep{zhao2023aliked}
                                                & 53.36 & \underline{89.71} & \underline{35.50} & \textbf{74.89} \\
Ours$^{\ast}$\,+\,LoFTR~\citep{sun2021loftr}
                                                & 52.69 & \textbf{90.31} & 34.48 & \underline{71.50} \\
Ours$^{\ast}$\,+\,EfficientLoFTR~\citep{wang2024eloftr}
                                                & \textbf{53.45} & 89.04 & \textbf{36.83} & 69.57 \\
Ours$^{\ast}$\,+\,RoMa~\citep{edstedt2024roma}  & 41.20 & 82.17 & 22.77 & 61.30 \\
\rowcolor{gray!12}
\textbf{Ours$^{\ast}$\,+\,\method (unified)}   & 52.41 & 88.38 & 34.26 & 70.80 \\
\bottomrule
\end{tabular}}\\[2pt]
\parbox{0.95\linewidth}{\centering\scriptsize Re-ranking
performed on Top-$K{=}5$ candidates with RANSAC
\textsc{usac\_magsac} at an $8$-pixel threshold, queries resized
to $322{\times}322$. \method remains within $1$ percentage point of the
separate DeDoDe teacher on every R@1 column while using one
shared DINOv2-B backbone forward per query.}
\end{table*}

\begin{table*}[!htbp]
\centering
\caption{End-to-end pose estimation on Urbanscape
($N_{\text{q}}{=}3{,}158$ test queries, $322{\times}322$,
Universal Sample Consensus and Marginalizing Sample Consensus scoring
(USAC-MAGSAC) at an $8$-pixel threshold). All pipelines use the Ours$^{\ast}$
retriever followed by the indicated matcher. ``Solver return'' is
the fraction of queries for which the perspective-$n$-point (PnP) solver
returns a finite pose; PnP Recall@1 (R@1) and Joint
$5^\circ\!/\!5\,$m R@1 columns are normalized over all
$N_{\text{q}}$ queries (PnP failures count as
mis-localizations). Median translation and rotation errors are
conditional diagnostics reported over all finite solver returns,
including finite but incorrect poses. \method uses one shared DINOv2-B
backbone forward per query; DeDoDe (separate) uses independent
DINOv2-B and DINOv2-L backbone forwards. Best and second-best results are bold and underlined,
respectively; tied best values are all bold.}
\label{tab:pnp-urbanscape}
\setlength{\tabcolsep}{3pt}
\scriptsize
\resizebox{\linewidth}{!}{%
\begin{tabular}{l|c|cc|ccc|c}
\toprule
& & \multicolumn{2}{c|}{PnP R@1, XY (\%)}
& \multicolumn{3}{c|}{Pose error (median, finite solver returns)}
& Joint $5^\circ\!/\!5\,$m \\
Pipeline
& Solver return (\%)
& $\tau=25$\,m & $\tau=50$\,m
& 3D (m) & XY (m) & Rotation ($^\circ$)
& R@1 (\%) \\
\midrule
Ours$^{\ast}$\,+\,DeDoDe (separate)~\citep{edstedt2024dedode}
                                               & \textbf{100.00} & 93.51 & 94.46 & 2.16 & 1.56 & 1.64 & 82.27 \\
Ours$^{\ast}$\,+\,ALIKED\,+\,LightGlue~\citep{zhao2023aliked,lindenberger2023lightglue}
                                               & 99.72 & 94.27 & 94.90 & 2.04 & 1.50 & 1.63 & 82.65 \\
Ours$^{\ast}$\,+\,ALIKED\,+\,MNN~\citep{zhao2023aliked}
                                               & 71.37  & 67.89 & 69.60 & 3.30 & 2.59 & 2.20 & 46.93 \\
Ours$^{\ast}$\,+\,LoFTR~\citep{sun2021loftr}
                                               & \textbf{100.00} & 94.46 & 94.84 & \underline{1.85} & \underline{1.36} & \underline{1.52} & \underline{84.61} \\
Ours$^{\ast}$\,+\,EfficientLoFTR~\citep{wang2024eloftr}
                                               & \underline{99.97} & \underline{94.49} & \underline{94.93} & 1.93 & 1.42 & 1.59 & 84.48 \\
Ours$^{\ast}$\,+\,RoMa~\citep{edstedt2024roma}
                                               & \textbf{100.00} & \textbf{95.38} & \textbf{95.79} & \textbf{1.47} & \textbf{0.97} & \textbf{1.31} & \textbf{88.95} \\
\rowcolor{gray!12}
\textbf{Ours$^{\ast}$\,+\,\method (unified)}
                                               & \textbf{100.00} & 93.00 & 94.02 & 2.29 & 1.70 & 1.77 & 79.16 \\
\bottomrule
\end{tabular}}
\end{table*}

The unified \method remains within $1$\,pp of the separate
DeDoDe teacher on every re-ranking column. It also retains a
$100\%$ PnP solver-return rate and reaches $93.00\%$ PnP Recall at
$25$\,m, compared with $93.51\%$ for the teacher. The stricter
joint score is $79.16\%$ versus $82.27\%$, exposing the modest
accuracy cost of replacing the dedicated DINOv2-L encoder with
the shared DINOv2-B representation. RoMa gives the best pose
accuracy under the reported joint metric and the lowest conditional
median errors, but it degrades retrieval ordering; its dense matcher is
roughly $6\times$ slower per probe pair than LoFTR, $20\times$
slower than EfficientLoFTR, and over two orders slower than
\method~(Table~\ref{tab:latency}). These results position \method as an
accuracy--efficiency design, not simply the most accurate
matcher.

\paragraph{Pose-solver robustness boundary}
The $100\%$ PnP solver-return rate in
Table~\ref{tab:pnp-urbanscape} suggests a potential advantage of
\method: its geometrically verified local matches provide a
solver-compatible correspondence distribution for every evaluated
Urbanscape query. Solver return is nevertheless distinct from
localization accuracy, because a finite PnP solution may still be
incorrect. Supplementary~\ref{app:pnp-stress} reports controlled
RGB-noise, central-occlusion, query-coordinate-noise, and
correspondence-corruption experiments to characterize the pipeline's
failure boundary and the associated reliability limitations.

\subsection{Latency and Efficiency}
\label{sec:exp-latency}

Latency is measured one query at a time on the RTX~3090 at
$322{\times}322$, Top-$K{=}5$, and 32-bit floating point (FP32). We average $100$
deterministic queries after three warm-up iterations and
synchronize every timed GPU region. Timings include query
encoding, gallery search, matching, device-to-host match-index
transfer, and CPU USAC-MAGSAC, while excluding common input/output (I/O),
resize, host-to-device transfer, cached database encoding, and the
standard final PnP solve. This protocol measures the online query
path through robust candidate verification rather than batched
throughput.

\begin{table*}[!htbp]
\centering
\caption{Per-query inference latency (ms) on Urbanscape at
$K{=}5$, single RTX 3090, $322{\times}322$, 32-bit floating point (FP32). \emph{One-time}
contains the query-side encoders and gallery search;
\emph{Top-5 matcher} contains the five GPU matching calls;
\emph{GPU subtotal} is their sum; and \emph{Top-5 CPU-USAC}
contains five host Universal Sample Consensus and Marginalizing Sample
Consensus (USAC-MAGSAC) calls. Numbers exclude input/output,
resizing, host-to-device transfer, cached database encoding, and the final
perspective-$n$-point solve. The 16-bit floating-point (FP16) numbers and
the full $K$-scaling breakdown are in
Table~\ref{tab:latency-scaling} of
Supplementary~\ref{app:latency-scaling}. Across all rows, best and second-best values are
bold and underlined, respectively; tied best values are all bold.}
\label{tab:latency}
\setlength{\tabcolsep}{3pt}
\scriptsize
\resizebox{\linewidth}{!}{%
\begin{tabular}{l|c|c|c|c|c}
\toprule
Pipeline (retriever\,+\,matcher)
   & \shortstack{One-time query\\+ retrieval GPU}
   & \shortstack{Top-5\\matcher GPU}
   & \shortstack{GPU\\subtotal}
   & \shortstack{Top-5\\CPU-USAC}
   & Total \\
\midrule
Ours$^{\ast}$\,+\,ALIKED\,+\,LightGlue~\citep{zhao2023aliked,lindenberger2023lightglue}
                                                    & 31.9 & 119.5 & 151.4 & \underline{9.5} & 160.9 \\
Ours$^{\ast}$\,+\,ALIKED\,+\,MNN~\citep{zhao2023aliked}
                                                    & 31.9 & \textbf{2.8} & \textbf{34.7} & \textbf{3.4} & \textbf{38.0} \\
Ours$^{\ast}$\,+\,LoFTR~\citep{sun2021loftr}
                                                    & \underline{21.4} & 361.9 & 383.3 & 27.4 & 410.7 \\
Ours$^{\ast}$\,+\,EfficientLoFTR~\citep{wang2024eloftr}
                                                    & 21.4 & 115.8 & 137.2 & 27.0 & 164.2 \\
Ours$^{\ast}$\,+\,DeDoDe (separate)~\citep{edstedt2024dedode}
                                                    & 53.9 & \underline{16.0} & 69.9 & 41.8 & 111.7 \\
Ours$^{\ast}$\,+\,RoMa~\citep{edstedt2024roma}
                                                    & \textbf{18.5} & 2325.5 & 2344.0 & 402.7 & 2746.7 \\
\midrule
\rowcolor{gray!12}
\textbf{Ours$^{\ast}$\,+\,\method (unified)}
                                                    & 38.2 & 16.0 & \underline{54.2} & 36.6 & \underline{90.8} \\
\bottomrule
\end{tabular}}\\[2pt]
\parbox{0.95\linewidth}{\scriptsize
The first three timing columns expose
$t_{\text{once,gpu}}$, $K\!\cdot\!t_{\text{match,gpu}}^{\text{pair}}$,
and their sum, respectively, following
Eqs.~\eqref{eq:rim-cost-baselines}--\eqref{eq:rim-cost-ours}.
For \method, $t_{\text{global}}$ is \emph{folded} into
$t_{\text{loc}}^{q}$ because the SALAD global descriptor and
the dense descriptor map are produced by the \emph{same}
DINOv2-B forward---this is the retrieval-in-matching
schedule's key saving. The \emph{CPU-USAC} column is
$K\!\cdot\!t_{\text{match,cpu}}^{\text{pair}}$ and is
implementation-agnostic; porting it to on-device GPU RANSAC
would shift this column towards the GPU side for \emph{every}
row without changing relative orderings. The ALIKED\,+\,MNN
row is a lightweight lower bound on latency but drops to a
$71.4\%$ solver-return rate and $46.9\%$ joint accuracy---see
Table~\ref{tab:pnp-urbanscape} and the discussion below;
every pipeline with a $100\%$ solver-return rate is strictly slower than
\method.}
\end{table*}

The measured online path through robust geometric verification
takes $90.8$\,ms per query: $1.2\times$ faster than
separate DeDoDe and $30.3\times$ faster than RoMa. Although
ALIKED+MNN is faster, its solver-return rate falls to $71.37\%$ and
its joint score to $46.93\%$; \method is therefore the fastest
evaluated pipeline with a $100\%$ solver-return rate. Its GPU-side saving follows the
design directly: the global descriptor and local descriptor map
reuse one DINOv2-B forward instead of separate global and local
backbones.

Full 16-bit floating-point (FP16) timings and $K\in\{5,10,20\}$ scaling are reported in
Supplementary~\ref{app:latency-scaling}. They preserve
the same ordering and confirm that \method and separate
DeDoDe share the shallow MNN slope, while dense pairwise
matchers scale with a much larger per-candidate forward.

\subsection{Component Ablation}
\label{sec:exp-ablation}

Table~\ref{tab:ablation-core} summarizes the contribution of
the two retriever-adaptation stages in the controlled run.

\begin{table}[!htbp]
\centering
\caption{Core two-stage ablation on Urbanscape. Recall@1 (R@1, \%) within
$50$\,m at $322{\times}322$ is reported for full 3D and planar XY
distances. The complete freeze-policy and
decoder-bridging ablations are reported in
Supplementary~\ref{app:component-ablations}.}
\label{tab:ablation-core}
\setlength{\tabcolsep}{6pt}
\footnotesize
\begin{tabular}{l|cc}
\toprule
Variant & R@1$_{50,3D}$ & R@1$_{50,XY}$ \\
\midrule
SALAD (no fine-tuning) & 53.93 & 76.79 \\
Stage 1                & 66.21 & 85.15 \\
\rowcolor{gray!12}
\textbf{Stage 1 + Stage 2} & \textbf{67.70} & \textbf{85.31} \\
\bottomrule
\end{tabular}
\end{table}

Stage~1 supplies most of the domain-transfer gain. Stage~2
adds $1.49$\,pp on altitude-sensitive 3D Recall while changing
the near-saturated planar score by only $0.16$\,pp, consistent
with its visually similar but geographically distant objective.
Supplementary~\ref{app:sample4geo-control} compares this one-shot
residual mining strategy with a matched Dynamic Similarity Sampling
protocol based on Sample4Geo~\citep{deuser2023sample4geo}.

\paragraph{Backbone control}
A matched DINOv3-B/16 extension~\citep{simeoni2025dinov3} raises
Urbanscape 3D R@1 at 25/50\,m from 9.44/33.44\% to 22.64/52.88\%
after Stage~1 and 22.83/53.83\% after Stage~2; geometric re-ranking
reaches 30.68/63.39\%, below DINOv2 \method\ at 34.26/70.80\%.
This control establishes feasibility after backbone-specific
adaptation, but does not imply a performance-neutral or guaranteed
improvement from replacing DINOv2 with DINOv3.
Supplementary~\ref{app:dinov3-rim-control} reports the standard-VPR,
two-stage, and decoder controls. Freeze-policy and
decoder-bridging results are retained in
Supplementary~\ref{app:component-ablations}.

\subsection{Discussion and Limitations}
\label{sec:exp-discussion}

\method trades at most $0.95$\,pp re-ranking Recall and
$3.11$\,pp strict joint pose Recall for $18.7\%$ lower
per-query latency ($90.8$ vs.\ $111.7$\,ms) relative to its
separate teacher.
Applications that prioritize submeter accuracy over latency
may prefer the decoupled model.

On Park, every evaluated pipeline remains below $10\%$ joint
$5$\,m/$5^\circ$ Recall. The near-nadir geometry weakens scale and
vertical-translation observability: rotation Recall remains high,
whereas position Recall at $5$\,m is the binding factor and translation
errors concentrate along the height direction. This pattern is
consistent with, but does not by itself prove, a scale/height
observability bottleneck. Park therefore remains a useful relative
zero-shot stress test; additional altitude priors or multiview evidence
are plausible directions for improving absolute pose accuracy.
Supplementary~\ref{app:extended-discussion},
\ref{app:park-gt}, and~\ref{app:park-altprior} provide the
complete bottleneck decomposition and sensor-prior analysis.

\section{Conclusion}
\label{sec:conclusion}

\paragraph{Recap}
We have presented \methodlong (\method), a UAV global visual
localization framework centered on shared-representation retrieval and
matching. A frozen global aggregation head and a distilled
local-descriptor decoder consume the same DINOv2-B token field, so one
shared DINOv2-B backbone forward per query supports both global
retrieval and local geometric re-ranking. UAV-viewpoint reference
sampling supplies a cross-view data mechanism, while two-stage
fine-tuning transfers appearance cues and suppresses visually similar
but geographically distant images. Together, these components form an
efficient inference system rather than a serial assembly of an existing
retriever and matcher.
Under the full 3D metric, \method lifts Recall@1 over SALAD by
$+8.55$/$+13.77$\,pp within $25$/$50$\,m on the reconstructed EPFL
urban benchmark and by $+4.45$/$+8.94$\,pp at the same radii on
Chang'an Park. It surpasses ten strong baseline families across all
eight Recall columns. At $K=5$, the measured online query path
through robust geometric verification takes $90.8$\,ms --- $1.2\times$ faster
than the strongest sparse-matching pipeline (DeDoDe, separate)
and over $30\times$ faster than SALAD\,+\,RoMa, at matched
re-ranking accuracy.

\paragraph{Limitations}
We identify four limitations of the present \method system:
\begin{itemize}
    \item \textbf{Geographic coverage of 3D reference maps.} The
    database is constructed from a 3D map and stores a pixel-aligned
    ECEF coordinate map for each rendered view. Google Photorealistic
    3D Tiles provide useful coverage mainly in urban areas, but can be
    sparse, low-resolution, or outdated in suburban, rural, and forested
    regions. Alternatively, reconstructing an SfM mesh requires a
    dedicated reference flight for each scene, as in Chang'an Park.
    These constraints limit the geographic scalability of the
    database-construction pipeline.
    \item \textbf{Reference-map temporal drift and open-set change.}
    The Park evaluation excludes 68 construction frames whose
    content is absent from the year-older reference model. The
    reported protocol therefore assumes that a valid counterpart
    exists in the database and does not establish robustness to
    newly built, demolished, or otherwise unmapped regions.
    \item \textbf{No explicit localization-confidence estimator.}
    The current pipeline returns its highest-ranked geometrically
    verified pose but does not calibrate localization confidence or
    provide a principled reject option. The corruption controls show
    that PnP can return a finite but incorrect pose after retrieval or
    correspondence quality has deteriorated; solver success alone is
    therefore not a reliability measure.
    \item \textbf{On-disk reference-database footprint.} Each
    reference view stores an RGB image, a pixel-aligned ECEF
    coordinate map, and camera extrinsics. The measured full
    databases occupy 80.6\,GiB on Urbanscape and 107.8\,GiB on Park.
    For on-disk storage on edge devices, however, the footprint
    remains substantial and will increase further as the reference
    database expands to larger geographic areas.
\end{itemize}

\paragraph{Future Work}
Each of the limitations above motivates a concrete direction:
\begin{itemize}
    \item \textbf{Pose estimation without stored per-pixel 3D coordinate maps.}
    Replace the per-pixel ECEF maps with query/reference depth
    prediction or a compact neural scene representation, such as a
    neural radiance field or 3D Gaussian Splatting model,
    from which candidate 3D points can be obtained on demand.
    \item \textbf{Database compression.} Investigate latent or
    quantized representations of both global descriptors and
    per-pixel ECEF coordinate maps, ideally optimized end-to-end with
    the matching loss, to reduce the on-disk database footprint while
    preserving localization accuracy.
    \item \textbf{Persistence-aware map anchoring under temporal drift.}
    Learn to favor stable structures such as road topology and
    established buildings, using semantic segmentation or visual
    attention to downweight construction activity, vehicles, and
    seasonal vegetation. Cross-time query--reference change detection
    should identify unsupported regions and trigger reference-map
    updating rather than forcing similarity to obsolete content.
    \item \textbf{Confidence-aware sequential integration.}
    Fuse retrieval margins, correspondence count and spatial coverage,
    geometric inlier statistics, PnP reprojection residuals, and
    temporal consistency into a calibrated reliability estimate.
    Accepted RIM poses can provide global corrections to visual
    odometry, visual--inertial odometry, or SLAM; rejected poses
    leave the sequential estimator active until RIM can reliably
    re-anchor the trajectory.
    \item \textbf{RIM component upgrades.}
    Extend RIM with stronger backbones such as DINOv3 and improved
    retrieval or matching modules through backbone-specific adaptation
    and distillation, while preserving one shared foundation-model
    backbone forward.
\end{itemize}

In summary, \method establishes feature sharing between global
retrieval and local description as an effective inference
paradigm for UAV localization. The same principle can extend
beyond the particular retriever, matcher, and 3D data source
used here.

% -------- KBS-required declarations (placed before references) --------
\section*{CRediT authorship contribution statement}
\textbf{Xin Li:} Conceptualization, Methodology, Software,
Validation, Formal analysis, Data curation,
Writing -- original draft, Writing -- review \& editing, Visualization.
\textbf{Siyuan Duan:} Software, Validation, Data curation,
Writing -- review \& editing.
\textbf{Shang Wang:} Investigation, Data curation,
Writing -- review \& editing.
\textbf{Zhimin Mao:} Investigation, Data curation.
\textbf{Bingliang Hu:} Resources, Supervision, Funding acquisition.
\textbf{Geng Zhang:} Supervision, Project administration,
Conceptualization, Writing -- review \& editing, Funding acquisition.

\section*{Declaration of competing interest}
The authors declare that they have no known competing financial
interests or personal relationships that could have appeared to
influence the work reported in this paper.

\section*{Data availability}
The real-UAV images underlying the Urbanscape benchmark are from
the publicly released EPFL dataset and remain available from the
original authors. Its Google Photorealistic 3D Tiles reference
renderings, together with the Google-derived Urbanscape-OOP
fine-tuning renderings, cannot be redistributed under the
applicable Google Maps Platform terms. The sampling parameters and
evaluation metadata needed to reproduce the protocol are provided
through the project repository, subject to each user's independently
authorized access and the applicable terms. Chang'an Park is
self-collected; access to its UAV queries, SfM-derived reference data,
pose metadata, and coordinate maps is documented in the project
repository. Project code, data-access instructions, and reproduction
instructions are publicly available at
\url{https://github.com/curious-energy/RIM}.

\section*{Funding}
This work was supported by the Open Research Fund of the Shaanxi
Key Laboratory of Optical Remote Sensing and Intelligent
Information Processing.

\section*{Declaration of generative AI and AI-assisted technologies in the writing process}
During the preparation of this work the authors used
OpenAI's GPT in order to improve the language, readability and
clarity of the manuscript. After using this tool, the authors
reviewed and edited the content as needed and take full
responsibility for the content of the publication.

% -------- bibliography --------
\bibliographystyle{rim-elsarticle-num-names}
\bibliography{refs}

% -------- supplementary material (after the references, in the same PDF) --------
\clearpage
\singlespacing
\typeout{RIM-ARXIV supplement: stretch=\baselinestretch; baseline=\the\baselineskip}
\section*{Supplementary Material}
\addcontentsline{toc}{section}{Supplementary Material}
\FloatBarrier
\appendix
\makeatletter
\@addtoreset{table}{section}
\@addtoreset{figure}{section}
\@addtoreset{equation}{section}
\@addtoreset{algorithm}{section}
\makeatother
\renewcommand{\thealgorithm}{\Alph{section}.\arabic{algorithm}}
% =============================================================
% Reproducibility detail moved from the KBS main manuscript.
% =============================================================
\section{Offline Hard-Negative Mining Procedure}
\label{app:hard-negative-algorithm}

The main paper describes the mining criterion in prose. For
implementation completeness, Algorithm~\ref{alg:offline-mining}
gives the original pseudocode.

\begin{algorithm}[H]
\captionsetup{font=footnotesize,skip=2pt}
\caption{Offline Statistical Hard-Negative Mining}
\label{alg:offline-mining}
\footnotesize
\begin{algorithmic}[1]
\setlength{\itemsep}{0pt}
\REQUIRE Stage-1 retriever $f_{\theta_1}$, database
         $\mathcal{D} = \{(I_i, \mathbf{T}_i)\}_{i=1}^N$,
         Top-$M$ cap, distance threshold $\tau_d$
\STATE $\mathbf{V} \gets [\,f_{\theta_1}(I_i)\,]_{i=1}^N$ \quad
       \COMMENT{precompute global descriptors}
\STATE $\mathcal{T} \gets \emptyset$
\FOR{each anchor $a \in \{1,\ldots,N\}$}
   \STATE $\mathcal{N}_a \gets \text{TopM}\bigl(
          \cos(\mathbf{V}_a, \mathbf{V}_{\cdot}),\, M\bigr)$
   \FOR{each $n \in \mathcal{N}_a$}
      \IF{$\|\mathbf{t}_a - \mathbf{t}_n\|_2 > \tau_d$
          \textbf{and} $\cos(\mathbf{V}_a,\mathbf{V}_n) > \tau_s$}
         \STATE pick a co-located positive $p$ with
                $\|\mathbf{t}_a - \mathbf{t}_p\|_2 < \tau_d/3$
         \STATE $\mathcal{T} \gets \mathcal{T} \cup \{(a, p, n)\}$
      \ENDIF
   \ENDFOR
\ENDFOR
\RETURN $\mathcal{T}$
\end{algorithmic}
\end{algorithm}

\FloatBarrier
% =============================================================
% Figures moved from the KBS main manuscript to reduce review length.
% =============================================================
\section{Additional Pipeline and Qualitative Results}
\label{app:additional-figures}

\begin{figure*}[!htbp]
\centering
\includegraphics[width=0.99\linewidth]{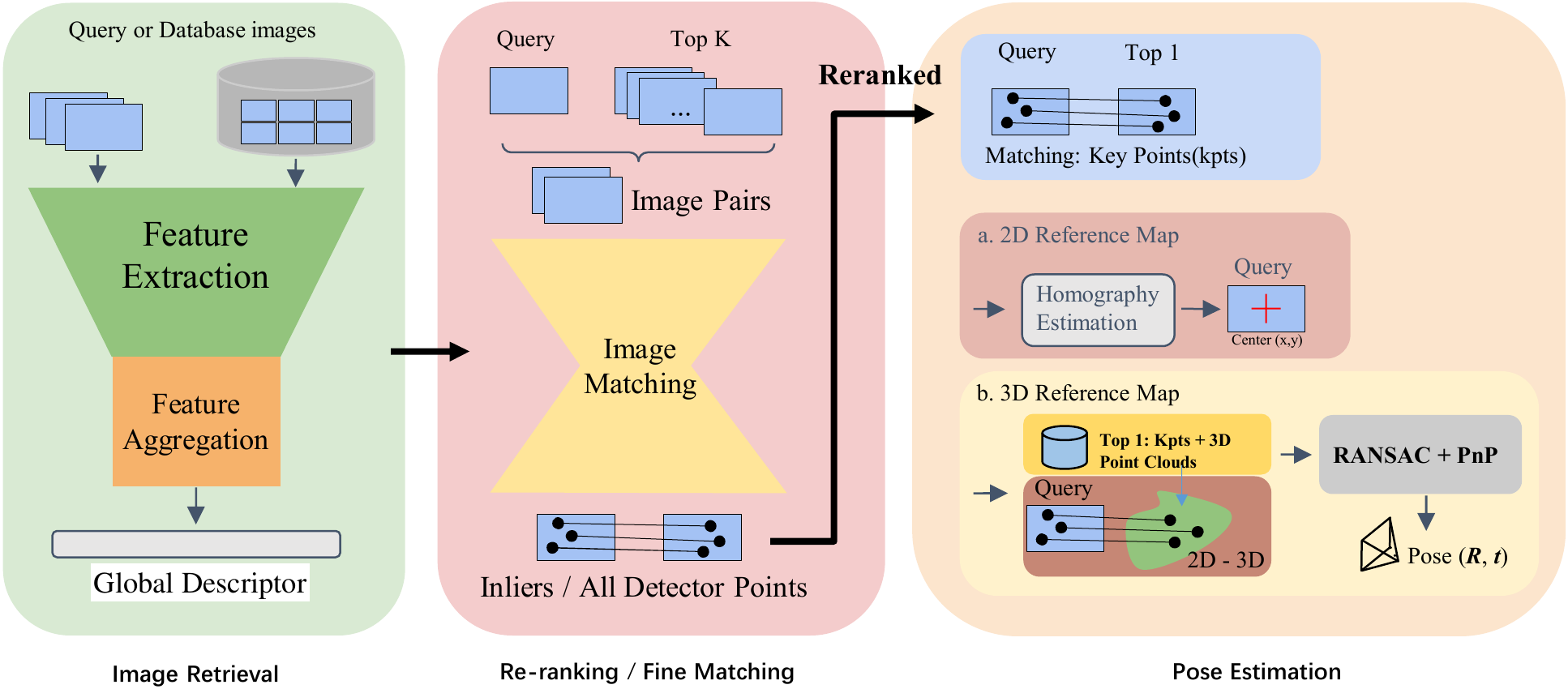}
\caption{\emph{Generic} retrieve-then-match-then-localize
pipeline for cross-domain UAV global visual localization, of
which virtually every prior system is an instance. The real
UAV query $I_q$ is encoded by a global retriever and matched
against the reference database
$\{I_i^{db}\}_{i=1}^{N}$ by cosine similarity to produce
Top-$K$ candidates. A separate local-feature branch extracts
descriptors for geometric re-ranking; database-side descriptors
can be cached for sparse pipelines, whereas dense pairwise
matchers run on every query--candidate pair. The resulting
2D--2D correspondences are verified by Universal Sample Consensus
and Marginalizing Sample Consensus (USAC-MAGSAC) fundamental-
matrix estimation, and the candidate with the largest
inlier set is promoted to Top-1. Because the database pixels
carry pixel-aligned Earth-Centered, Earth-Fixed (ECEF) coordinates
(Section~\ref{sec:method-database}), the Top-1 2D--2D inlier
set is lifted, in a single indexing step, to a 2D--3D
correspondence set, and the final six-degree-of-freedom (6-DoF) UAV pose is solved by
random sample consensus (RANSAC) with perspective-$n$-point (PnP) on
the full ECEF coordinates in double precision; the
estimated camera center is then expressed relative to the area origin
(Section~\ref{sec:method-pose}). A conventional sparse pipeline
therefore uses separate global and local query encoders; a dense
pairwise pipeline additionally repeats its matching network over
the $K$ candidates. Section~\ref{sec:method-rim} and
Fig.~\ref{fig:rim-framework} show how \method removes the
separate global DINOv2 query encoding by sharing one DINOv2-B
representation between the two branches.}
\label{fig:pipeline}
\end{figure*}

\begin{figure}[!htbp]
\centering
\includegraphics[width=0.68\linewidth]{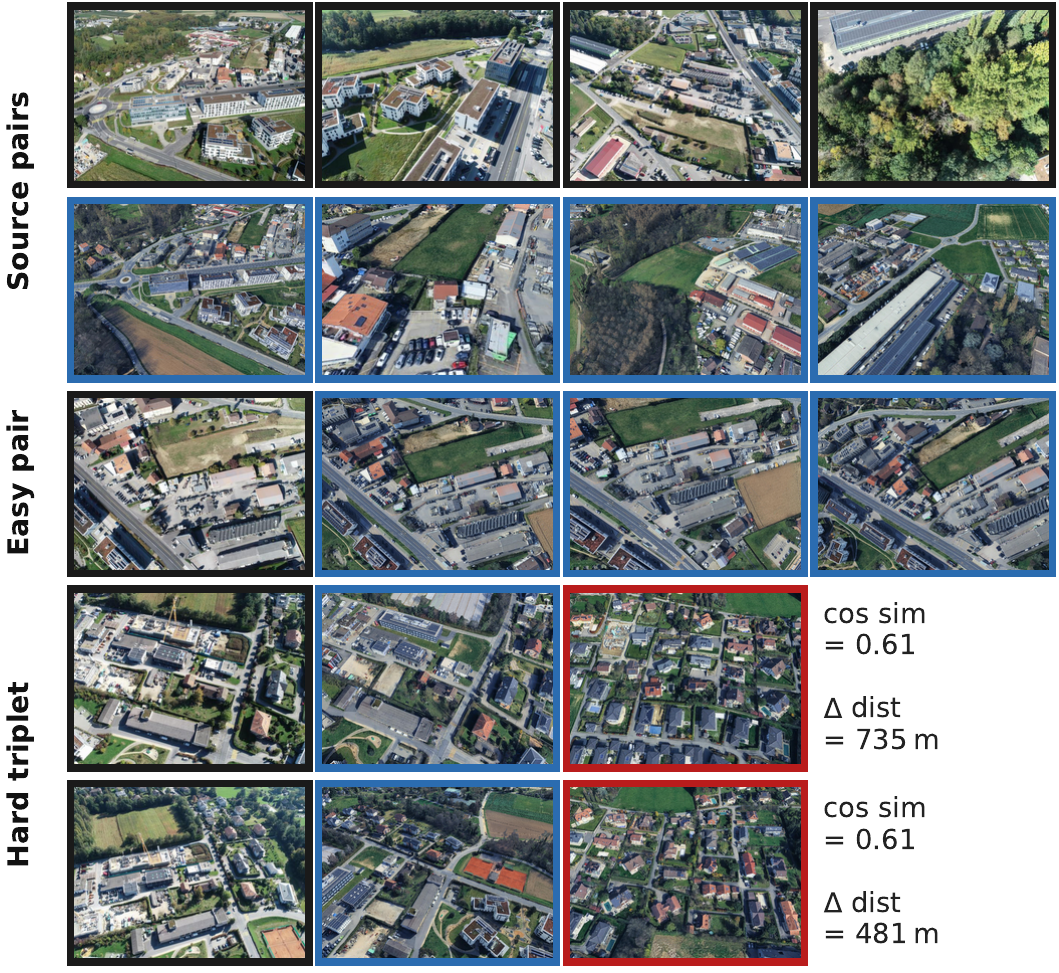}
\caption{Concrete samples drawn from the two-stage
fine-tuning dataset (Urbanscape Out-of-Place, Urbanscape-OOP; geographically disjoint
from both test benchmarks; see Table~\ref{tab:dataset-stats}).
Left-edge colours encode the image source: black\,=\,real UAV,
blue\,=\,3D-Tiles rendering, red\,=\,mined hard negative.
\emph{Source pairs:} four real UAV queries paired with the
nearest UAV-perspective renderings sampled from a
photogrammetric 3D mesh via the grid protocol of
Section~\ref{sec:method-database} -- the raw cross-domain
material the fine-tuner has to align.
\emph{Easy pair:} one real UAV anchor followed by three
renderings under bounded translation and orientation
perturbations, kept as pose-near positives by
the multi-similarity loss of Strategy~1.
\emph{Hard triplet:} two mined hard-negative triplets
[anchor\,$|$\,positive\,$|$\,hard-negative]. The number reported
above each triplet is the anchor-to-hard-negative cosine
similarity, and $\Delta$ distance is the verified Earth-Centered,
Earth-Fixed (ECEF) distance between the positive and the
hard-negative. Both triplets are nearly indistinguishable to a
visual retriever (cosine similarity $\sim\!0.6$ to the anchor)
yet correspond to grid tiles separated by several hundred metres
on the ground -- exactly the failure mode Strategy~2 is
designed to fix.}
\label{fig:finetune-samples}
\end{figure}

\begin{figure*}[!htbp]
\centering
\begin{subfigure}{0.49\linewidth}
  \centering
  \includegraphics[width=\linewidth]{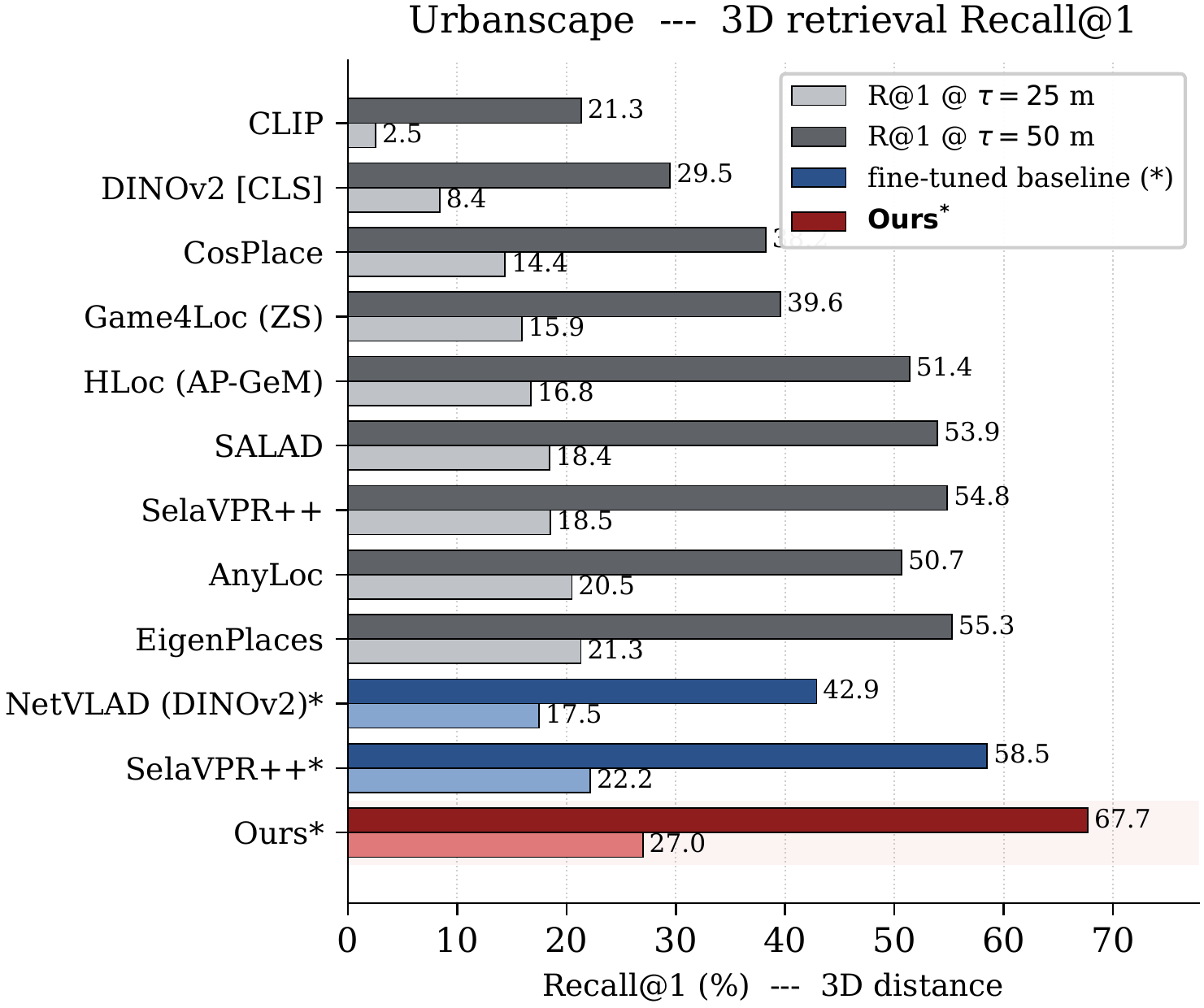}
  \caption{Urbanscape (EPFL reconstructed).}
  \label{fig:rec-bars-urb}
\end{subfigure}\hfill
\begin{subfigure}{0.49\linewidth}
  \centering
  \includegraphics[width=\linewidth]{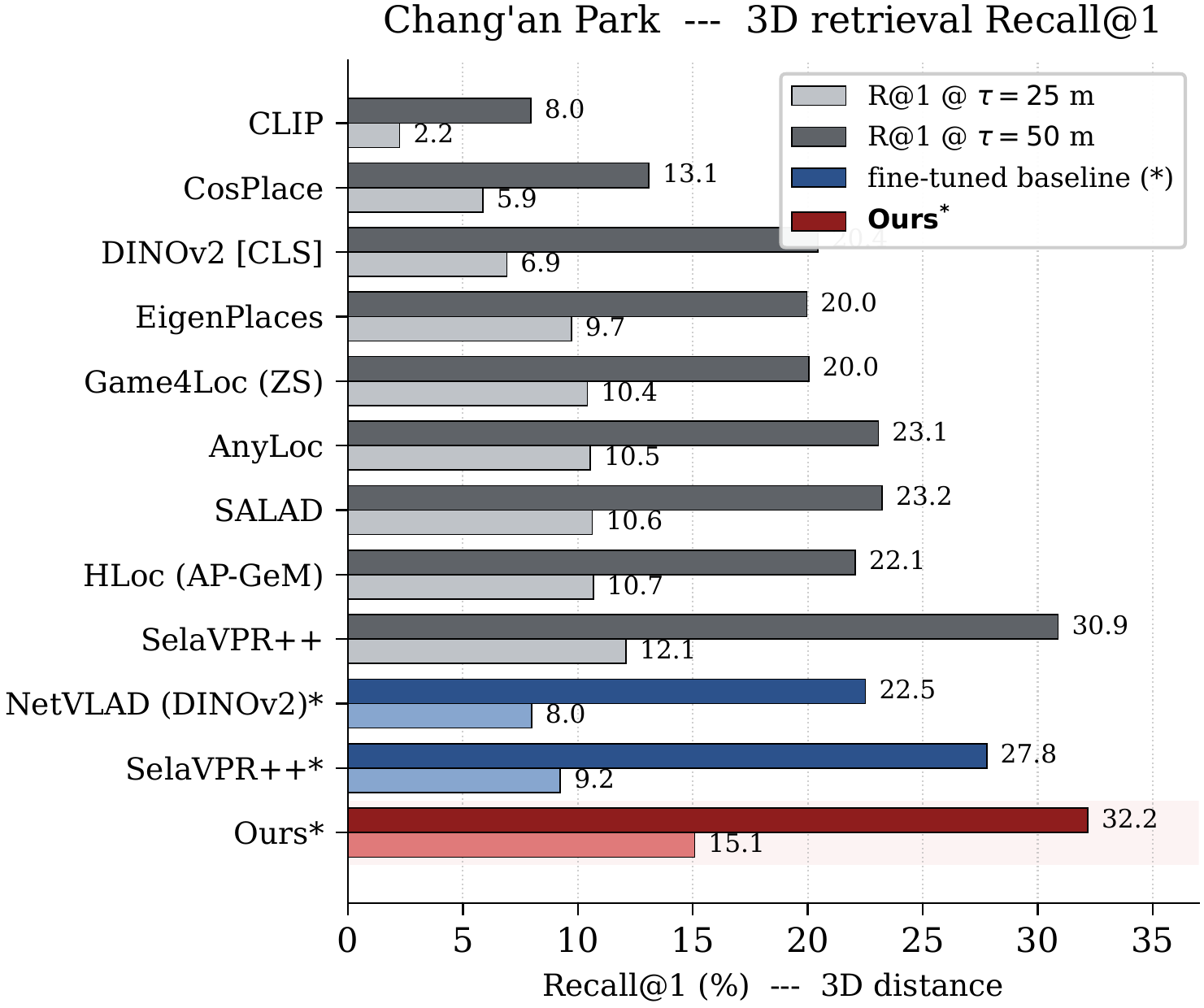}
  \caption{Chang'an Park (self-collected).}
  \label{fig:rec-bars-park}
\end{subfigure}
\caption{3D Recall@1 at $\tau{=}25$\,m and $\tau{=}50$\,m for
all competitors on the two benchmarks, plotted directly from
Tables~\ref{tab:rec-urbanscape-3d} and~\ref{tab:rec-park-3d}.
Ours$^{\ast}$ (red) tops every competitor at both plotted
tolerances on both benchmarks. NetVLAD(DINOv2)$^{\ast}$ is a
matched-data fine-tuning baseline, while SelaVPR++ is shown in
both released and same-data fine-tuned forms. Neither control
reproduces the consistent gains of Ours$^{\ast}$, supporting an
interaction between the SALAD aggregation head and the
two-stage recipe rather than a generic benefit from exposure
to the paired dataset.}
\label{fig:rec-bars}
\end{figure*}

\begin{figure*}[!htbp]
\centering
\includegraphics[width=0.99\linewidth]{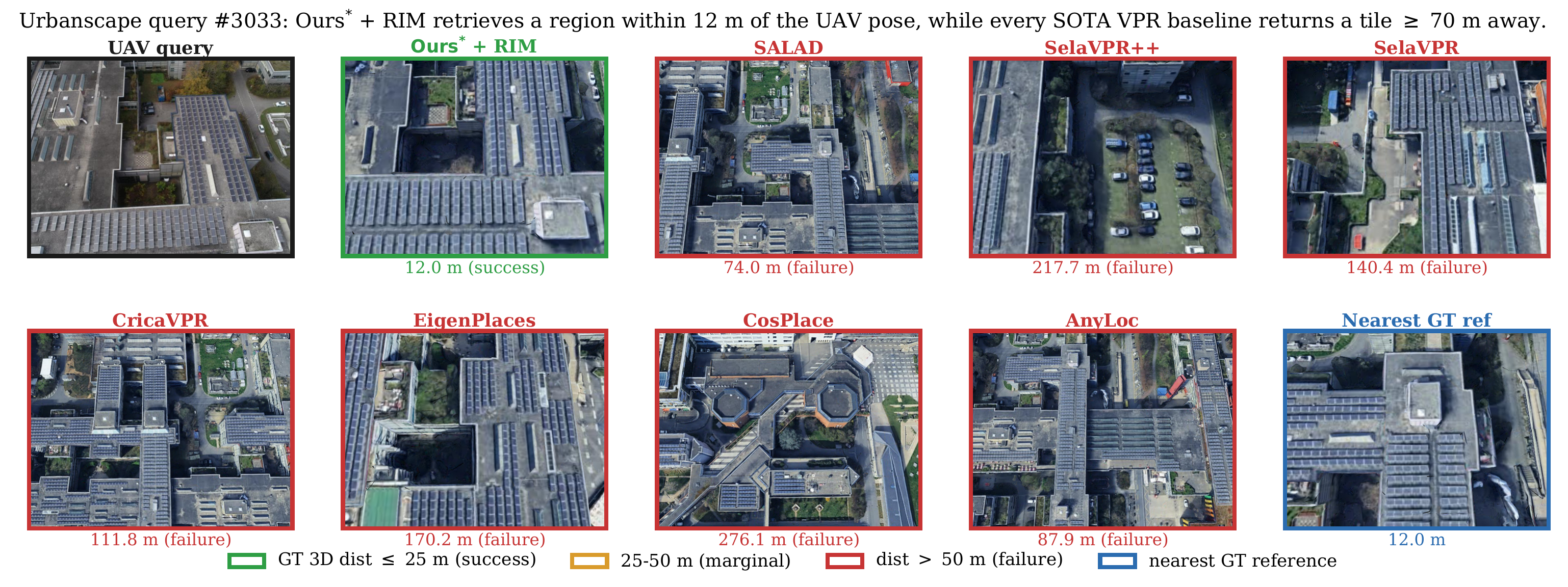}
\caption{Qualitative cross-domain retrieval on Urbanscape query
\#3033 (black border). The remaining panels show the Top-1
reference tile returned by Ours$^{\ast}$\,+\,\method and by
seven state-of-the-art (SOTA) visual-place-recognition (VPR)
baselines run on the \emph{same} query at the
\emph{same} input resolution ($322{\times}322$): the five
methods that also appear in the main recall tables --- SALAD
\citep{izquierdo2024salad}, SelaVPR$^{++}$
\citep{lu2025selavprpp}, EigenPlaces
\citep{berton2023eigenplaces}, CosPlace
\citep{berton2022cosplace}, AnyLoc
\citep{keetha2023anyloc} --- plus two additional contemporary
VPR baselines that we evaluated on the same cache for
robustness, SelaVPR (T-PAMI'24) and CricaVPR (CVPR'24).
Ours$^{\ast}$\,+\,\method retrieves a tile $12$\,m from the
true UAV pose (green border, success), while \emph{every} SOTA
baseline returns a tile $\geq 74$\,m away (red borders,
failures spanning $74$--$276$\,m). The bottom-right panel is
the nearest 3D-Tiles ground-truth reference for context. All
distances are 3D Euclidean in area-centred ECEF coordinates;
For the methods shared with the main quantitative comparison,
full-test ranking statistics are reported in
Tables~\ref{tab:rec-urbanscape-3d} and
\ref{tab:rec-park-3d} and visualised in
Fig.~\ref{fig:rec-bars}; SelaVPR and CricaVPR are included here
as additional qualitative controls.}
\label{fig:sota-visual}
\end{figure*}

\begin{figure*}[!htbp]
\centering
\includegraphics[width=0.72\linewidth]{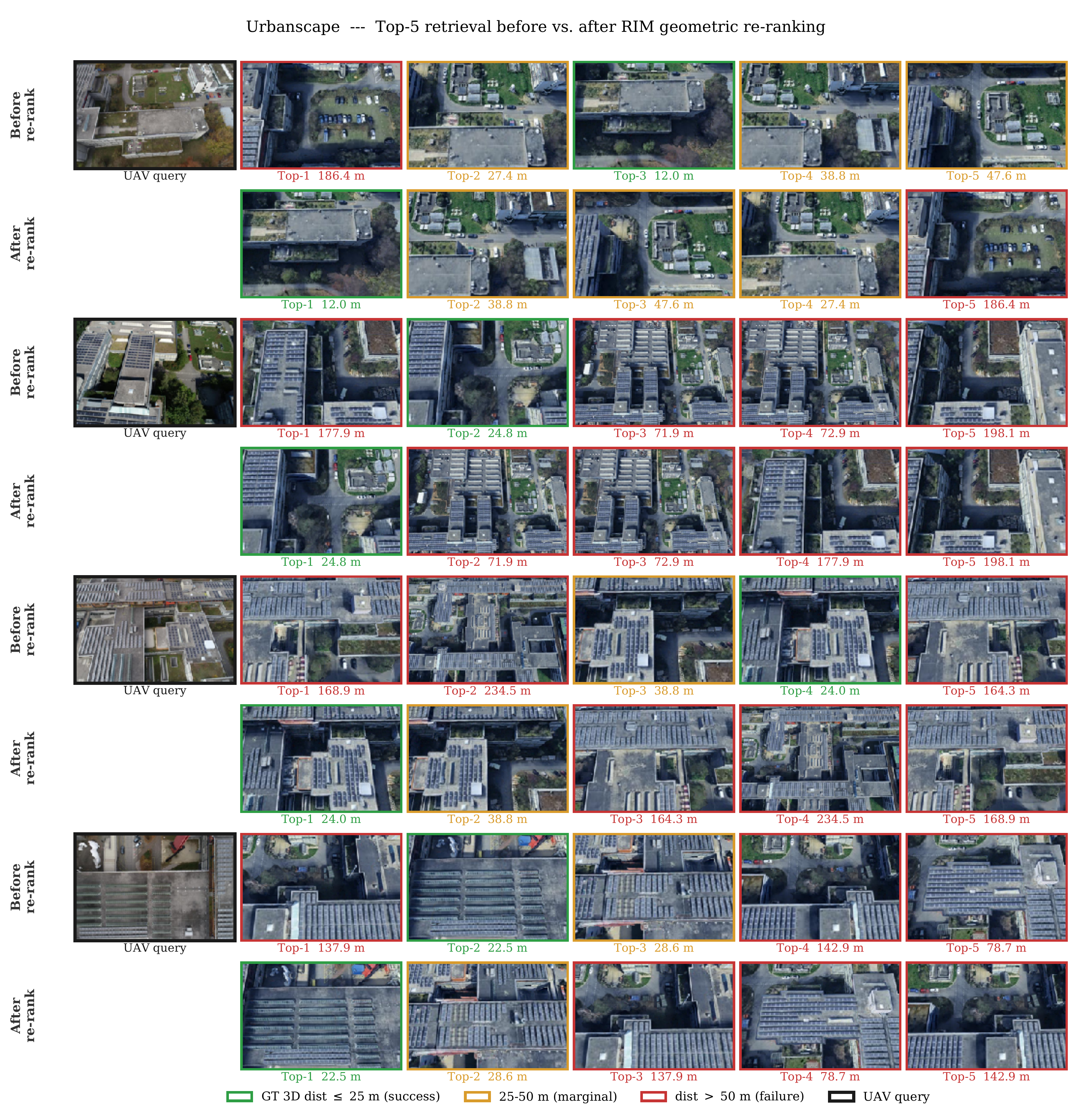}
\caption{Qualitative effect of geometric re-ranking on
Urbanscape. Each case shows the same query (leftmost column)
with its ``before re-rank'' Top-$K$ (upper row) -- the
retrieval-only order returned by Ours$^{\ast}$ -- and the
``after re-rank'' Top-$K$ (lower row) produced by the
robust fundamental-matrix verifier that feeds on the
\method dense local descriptors. Green borders mark references
within $25$\,m of the ground-truth UAV position, red borders
mark references farther than $50$\,m. The top retrieval
candidate is frequently a visually plausible but
geometrically-off tile; once the \method re-ranker runs, the
correct tile is promoted to the first position in every case,
which is what ultimately feeds perspective-$n$-point (PnP) estimation
and produces the pose
Recall gains reported in Table~\ref{tab:pnp-urbanscape}.}
\label{fig:rerank-qual}
\end{figure*}

Figure~\ref{fig:pipeline} expands the generic localization
pipeline, and Fig.~\ref{fig:finetune-samples} shows concrete
fine-tuning samples. Figures~\ref{fig:rec-bars} and
\ref{fig:sota-visual} provide aggregate and qualitative retrieval
comparisons, while Fig.~\ref{fig:rerank-qual} visualizes the effect
of geometric re-ranking.

\FloatBarrier
% =============================================================
% Detailed experiments moved from the KBS main manuscript.
% The pre-compression manuscript is preserved in the dated archive.
% =============================================================
\section{Extended Component Ablations}
\label{app:component-ablations}

\paragraph{Stage-1 freeze policy}
Table~\ref{tab:ablation-freeze} compares four backbone-update
policies for Stage 1, all evaluated under the identical
$322{\times}322$ retrieval protocol of the main tables:
\emph{no freeze} (everything updated), \emph{frozen backbone}
(the complete DINOv2-B backbone and global-token projection frozen,
our final choice), an \emph{original DINOv2} initialisation without
SALAD weights, and the original SALAD checkpoint without
fine-tuning. Two findings guide the design. First, the
SALAD initialisation matters: dropping the SALAD head
entirely (row~2) collapses R@1$_{50,3D}$ to $39.3\%$, nearly
$15$\,pp below the vanilla SALAD checkpoint and more than
$25$\,pp below either fine-tuned variant. Second, among the
fine-tuned Stage-1 variants the frozen-backbone policy is the
numerical leader on both metrics (R@1$_{50,3D}=66.21\%$,
R@1$_{50,XY}=85.15\%$), but its advantage over the
fully-updated variant is only $0.12$\,pp on 3D R@1 and
$0.19$\,pp on XY R@1, both below the single-run sampling
uncertainty at our query budget ($N=3158$ real UAV queries).
We therefore adopt the frozen-backbone setting because it is the
numerical leader and prevents representation drift on the small UAV
training set, while still letting the soft-cluster, score, and dust-bin
aggregation parameters absorb the
UAV aliasing patterns on the small-vehicle and dense-foliage
tiles that Stage~2 subsequently hard-mines.

\begin{table}[!htbp]
\centering
\caption{Stage-1 backbone freeze policy ablation. Recall@1 (R@1, \%)
within $50$\,m on Urbanscape (3D and XY metrics,
$322{\times}322$, one-epoch budget). Bold: best per column;
shaded row: \method's adopted configuration. Freezing the complete
DINOv2-B backbone and global-token projection is the numerical winner,
but the margin over the fully-updated variant
($+0.12$ percentage points in 3D and $+0.19$ percentage points in XY) is below the sampling
uncertainty at $N=3158$; the practical motivation is to preserve
the foundation-model representation while adapting only SALAD's
aggregation parameters.}
\label{tab:ablation-freeze}
\setlength{\tabcolsep}{6pt}
\footnotesize
\begin{tabular}{l|cc}
\toprule
Variant & R@1$_{50,3D}$ & R@1$_{50,XY}$ \\
\midrule
SALAD (no fine-tune)              & 53.93 & 76.79 \\
DINOv2 init (no SALAD weights)    & 39.27 & 56.24 \\
Stage-1 (no freeze, all updated)  & 66.09 & 84.96 \\
\rowcolor{gray!12}
\textbf{Stage-1 (frozen backbone, Ours)} & \textbf{66.21} & \textbf{85.15} \\
\bottomrule
\end{tabular}
\end{table}

\paragraph{Stage-2 contribution}
Table~\ref{tab:ablation-stage2} isolates the contribution of
the hard-negative refinement stage, continuing from the
Stage-1 frozen-backbone checkpoint of
Table~\ref{tab:ablation-freeze} and evaluated under the same
$322{\times}322$ protocol. The historical Stage~2 run uses a sparse
strict-negative pool built from the \emph{geographically} distant but
\emph{visually} similar tile subset described in
Section~\ref{sec:method-finetune}, yet still lifts
R@1$_{50,3D}$ from the Stage-1 baseline of $66.21\%$ to
$67.70\%$ (a ${+}1.49$\,pp gain), while the XY-projected
Recall moves from $85.15\%$ to $85.31\%$ (${+}0.16$\,pp).
The asymmetry between the two axes is the scientifically
interesting signal here. After Stage 1 the lateral retrieval
metric is already close to saturation at this tolerance
($85.15\%$ XY R@1 within $50$\,m), so the room Stage~2 has to
move it is intrinsically narrow and the observed
${+}0.16$\,pp lift is below the sampling uncertainty of our query
budget ($N=3158$), so this experiment
does not resolve an XY gain or loss. The 3D metric, by contrast,
additionally penalises altitude error, and on this axis
Stage~2 delivers a $+1.49$\,pp gain --- a factor of
$9{\times}$ larger than the XY movement. This pattern is
consistent with Stage~2 correcting ambiguities that are exposed
by the altitude-sensitive 3D metric, because its mined negatives
use 3D geographic separation and can include visually similar
views at different camera heights. Aggregate Recall alone does
not identify the corrected error axis causally, however; doing so
would require paired per-query error transitions and altitude-
stratified analysis. We therefore treat altitude-aware correction
as a supported interpretation rather than a demonstrated
mechanism.

\begin{table}[!htbp]
\centering
\caption{Stage-2 hard-negative refinement ablation, continuing
from the Stage-1 freeze-global row of
Table~\ref{tab:ablation-freeze}. Stage~2 lifts the
altitude-sensitive 3D Recall@1 by a larger margin
while leaving the near-saturated planar (XY) Recall@1 essentially unchanged
---a pattern consistent with, but not causal proof of, correction
of altitude-sensitive ambiguities. Recall@1 (R@1, \%)
within $50$\,m on Urbanscape, $322{\times}322$.}
\label{tab:ablation-stage2}
\setlength{\tabcolsep}{6pt}
\footnotesize
\begin{tabular}{l|cc}
\toprule
Variant & R@1$_{50,3D}$ & R@1$_{50,XY}$ \\
\midrule
Stage 1 (frozen backbone, best)       & 66.21 & 85.15 \\
\rowcolor{gray!12}
\textbf{Stage 1 + Stage 2 (Ours)}     & \textbf{67.70} & \textbf{85.31} \\
$\Delta$                              & $+1.49$ & $+0.16$ \\
\bottomrule
\end{tabular}
\end{table}

\paragraph{Decoder--backbone bridging strategy}
The student backbone (DINOv2-B) emits $768$-channel tokens,
whereas the original DeDoDe descriptor decoder consumes
$1024$ channels at its top stage. There are three natural
ways to bridge this width gap when distilling
(Section~\ref{sec:method-rim}): (i)~\textbf{native-$768$} ---
re-instantiate the decoder's top stage at $768$\,channels and
train the entire decoder from scratch under the
\method distillation loss, the choice we adopt in the paper;
(ii)~\textbf{multilayer-perceptron (MLP) adapter} --- prepend a two-layer MLP
($768 \to 1024$) and re-use the pre-trained DeDoDe decoder
weights; or (iii)~\textbf{Conv adapter} --- replace the MLP
with a $1{\times}1$ convolutional adapter (same width
remapping, fewer parameters). The three variants share the
SALAD retriever, the DeDoDe detector and the RANSAC-PnP
solver of Sections~\ref{sec:method-rim}--\ref{sec:method-pose};
they only differ in the descriptor decoder. We additionally
report the un-distilled DeDoDe-G teacher (DINOv2-L+VGG19,
$1024$-dim, no \method distillation) as the teacher reference
that the distilled $768$-channel students are trained to
approach. Table~\ref{tab:ablation-adapter} compares the three
bridging choices with this reference.

\begin{table}[!htbp]
\centering
\caption{Decoder bridging ablation between the DINOv2-B
($768$\,d) backbone and the DeDoDe descriptor decoder
($1024$\,d). All four variants share the same Ours$^{\ast}$
retriever, Universal Sample Consensus and Marginalizing Sample Consensus
(USAC-MAGSAC) verifier, and perspective-$n$-point (PnP) solver, evaluated on
the $3{,}158$-query Urbanscape test split at $322{\times}322$.
``DeDoDe-G (no distill)'' is the un-distilled teacher reference.
The native-$768$ row is \method's
adopted design.}
\label{tab:ablation-adapter}
\setlength{\tabcolsep}{3pt}
\scriptsize
\resizebox{\linewidth}{!}{%
\begin{tabular}{l|cc|cc|ccc|c}
\toprule
& \multicolumn{2}{c|}{XY R@1 (\%)}
& \multicolumn{2}{c|}{3D R@1 (\%)}
& \multicolumn{3}{c|}{PnP error (med)}
& Joint \\
Decoder variant & $\tau{=}25$ & $\tau{=}50$ & $\tau{=}25$ & $\tau{=}50$
& 3D (m) & XY (m) & Rotation ($^\circ$) & $5^\circ\!/5\,$m \\
\midrule
DeDoDe-G (no distill, teacher)                & 53.58 & 88.79 & 35.28 & 71.41 & \textbf{2.13} & \textbf{1.60} & \textbf{1.71} & \textbf{81.10} \\
\midrule
\rowcolor{gray!12}
\textbf{Native-$768$ (Ours)}                  & 52.41 & 88.38 & 34.26 & 70.80 & 2.29 & 1.70 & 1.77 & \textbf{79.16} \\
MLP adapter ($768{\to}1024$)                  & 53.26 & 88.73 & \textbf{35.31} & 70.84 & 2.33 & 1.73 & 1.76 & 78.06 \\
Conv adapter ($1{\times}1$, $768{\to}1024$)   & 53.26 & 87.90 & 34.86 & 70.52 & 2.36 & 1.78 & 1.84 & 77.33 \\
\bottomrule
\end{tabular}}\\[2pt]
\parbox{0.97\linewidth}{\centering\scriptsize
PnP error medians are reported over finite solver returns
(solver-return rate $= 100\%$ on every row). \textbf{Bold} marks the
best value in each column. Among the three distilled students
the differences are below $1.5$\,pp on every column except
Joint $5^\circ\!/\!5\,$m, on which native-$768$ leads the next
best (MLP adapter) by $1.10$\,pp and the worst (Conv adapter)
by $1.83$\,pp. All three distilled students are within
$3$--$4$\,pp of the un-distilled teacher's Joint
$5^\circ\!/\!5\,$m, in line with the other re-ranking and PnP
columns. The simplest bridging choice---widen the decoder's
top stage to $768$ and train it from scratch---is therefore
also the best on the end-to-end pose metric, which is why we
adopt it in \method.}
\end{table}

Three observations are worth highlighting.
First, the three distilled students are tightly clustered:
their pairwise gaps are below $1.5$\,pp on every retrieval and
median-error column, and the solver-return rate is identical
($100\%$) across the four rows. This indicates that, given a
fixed teacher and a fixed distillation loss, the decoder's
input-bridging mechanism is a second-order design choice for
overall coverage --- the heavy lifting is done by the
shared distilled backbone.
Second, the simplest bridging choice (native-$768$, no
adapter) is also the best on the strictest metric, Joint
$5^\circ\!/5\,$m R@1, leading the MLP adapter by $1.10$\,pp
and the Conv adapter by $1.83$\,pp; it also produces the
highest mean PnP inlier count among the three students
($295$ versus $276$ and $275$). The intuition is consistent
with the distillation set-up of
Section~\ref{sec:method-rim}: re-using the pre-trained DeDoDe
decoder weights through a learned width-bridging adapter
preserves the teacher's $1024$-d topology but pays an extra
optimisation cost (the adapter must absorb the entire
DINOv2-L$\to$DINOv2-B distribution shift in a thin layer),
which limits how faithfully the student's local geometry
matches the teacher's. Re-instantiating the top stage at
$768$ and training the decoder end-to-end avoids that
bottleneck.
Third, the un-distilled DeDoDe-G teacher (top row) leads on
median position and rotation errors and on Joint
$5^\circ\!/5\,$m, but by no more than $1.94$\,pp on the reported
metrics relative to native-$768$. This confirms that the distilled
$768$-channel student approaches the $1024$-channel teacher
without exceeding it, while in exchange enabling the
$1.2\times$ measured online-path speed-up visible in
Table~\ref{tab:latency} (\method \emph{vs.}
DeDoDe, separate)---the central trade-off promised by
Section~\ref{sec:method-rim}.

\subsection{Matched DINOv3 Extension}
\label{app:dinov3-rim-control}

Because both learned interfaces in \method\ were trained in the DINOv2
representation space, a matched cross-family evaluation requires adapting
both the retrieval and local-descriptor interfaces. The SALAD aggregation
branch consumes DINOv2 patch/global-token statistics, and the local decoder
was distilled from a DINOv2-L/VGG19 DeDoDe-G teacher. We therefore
constructed a complete DINOv3 extension. Its global branch is the audited
DINOv3-B/16~\citep{simeoni2025dinov3} Stage-2 SALAD retriever; its local branch combines the same
frozen DINOv3 token map with the frozen DeDoDe VGG19 features. One
DINOv3 forward consequently supplies both the 8,448-dimensional global
descriptor and the token field used to form the 256-channel dense local
descriptor map.

\paragraph{Retrieval basis and two-stage adaptation}
We first isolate the quality of the retrieval component before testing
the shared local branch. Table~\ref{tab:dinov3-standard-vpr} compares
the released DINOv2-B/14 SALAD checkpoint with three DINOv3-B/16 SALAD
adaptation strengths on the standard MSLS-val and Pitts30k protocols.
The DINOv3 variants were trained on the complete official 23-city
GSV-Cities subset; for each dataset and configuration, the checkpoint is
selected by R@1 from its complete validation curve, and its corresponding
R@5/R@10 are reported without per-metric checkpoint selection. DINOv3
benefits from backbone adaptation, but even the best DINOv3 setting trails
released DINOv2 SALAD by 6.49\,pp on MSLS-val and 3.18\,pp on Pitts30k
R@1. We therefore use the last-two-block initialization for the matched
Urbanscape control because it performs better on Pitts30k.

\begin{table}[!htbp]
\centering
\caption{Retrieval basis before UAV-domain adaptation. Recall@$K$ (\%)
on standard visual-place-recognition datasets. Each DINOv3 configuration is selected by Recall@1 (R@1)
independently on each dataset; R@5/R@10 come from that same checkpoint.
Best values are in bold and second-best values are underlined.}
\label{tab:dinov3-standard-vpr}
\setlength{\tabcolsep}{3.2pt}
\scriptsize
\resizebox{\linewidth}{!}{%
\begin{tabular}{l|ccc|ccc}
\toprule
& \multicolumn{3}{c|}{MSLS-val} & \multicolumn{3}{c}{Pitts30k} \\
Retrieval initialization & R@1 & R@5 & R@10 & R@1 & R@5 & R@10 \\
\midrule
DINOv3-B/16, frozen backbone
  & 77.97 & 89.32 & 91.89 & 87.68 & 94.23 & 95.88 \\
DINOv3-B/16, last two blocks trainable
  & 84.05 & 93.24 & \underline{94.86} & \underline{89.22} & \underline{95.19} & \underline{96.82} \\
DINOv3-B/16, last four blocks trainable
  & \underline{85.54} & \underline{93.51} & 94.73 & 88.57 & 95.03 & 96.64 \\
Released DINOv2-B/14 SALAD
  & \textbf{92.03} & \textbf{96.35} & \textbf{96.89}
  & \textbf{92.40} & \textbf{96.29} & \textbf{97.36} \\
\bottomrule
\end{tabular}}
\end{table}

Table~\ref{tab:dinov3-two-stage} then follows the selected DINOv3
retriever through the complete Urbanscape adaptation path. The
pre-adaptation checkpoint uses the GSV-Cities-trained last-two-block
initialization. Stage~1 freezes DINOv3 and its global-token projection
and updates only 885,953 SALAD cluster, score, and dust-bin parameters
on reconstructed positive pairs. It supplies the dominant gain:
+13.20/+18.52\,pp at 25\,m and +19.44/+24.86\,pp at 50\,m for
R@1/R@5 (all deltas are computed before display-value rounding).

\begin{table}[!htbp]
\centering
\caption{DINOv3 retrieval adaptation on all 3,158 Urbanscape queries.
Values are 3D Recall@$K$ (\%). Best values are in bold and second-best
values are underlined.}
\label{tab:dinov3-two-stage}
\setlength{\tabcolsep}{4pt}
\footnotesize
\resizebox{\linewidth}{!}{%
\begin{tabular}{l|cc|cc}
\toprule
& \multicolumn{2}{c|}{$25$\,m} & \multicolumn{2}{c}{$50$\,m} \\
Checkpoint & R@1 & R@5 & R@1 & R@5 \\
\midrule
Last-two-block initialization & 9.44 & 17.16 & 33.44 & 53.58 \\
+ Stage~1 positive-pair adaptation
  & \underline{22.64} & \underline{35.69}
  & \underline{52.88} & \underline{78.44} \\
+ Stage~2 hard-negative refinement
  & \textbf{22.83} & \textbf{36.00}
  & \textbf{53.83} & \textbf{78.63} \\
\bottomrule
\end{tabular}}
\end{table}

Stage~2 further improves all four retrieval metrics. The final endpoint
gives total gains over its pre-adaptation initialization of
+13.39/+18.84\,pp at 25\,m and +20.39/+25.05\,pp at 50\,m.
The comparison also separates backbone quality from adaptation gain:
the DINOv3 initialization is weak under the cross-domain 3D metric, but
the two-stage UAV adaptation recovers a substantial part of that gap.

For reference, the adopted DINOv2 path exhibits the same division of
labour under the paper protocol. Released SALAD gives 53.93\% 3D
R@1 at 50\,m; Stage~1 raises it to 66.21\% (+12.28\,pp), and the
sparse Stage~2 refinement reaches 67.70\% (+1.49\,pp over Stage~1),
as isolated in Tables~\ref{tab:ablation-freeze} and
\ref{tab:ablation-stage2}. Its final Top-5 shared geometric re-ranking
reaches 70.80\%. Thus, for both backbone families, positive-pair
camera/domain alignment provides most of the retrieval improvement,
whereas Stage~2 is a smaller residual-confusion correction whose mining
rule must be calibrated to the representation family.

For local cross-family adaptation, only the 5.59\,M-parameter decoder,
initialized from the DINOv2-trained \method\ decoder, is updated; the
DINOv3 backbone, SALAD head, VGG19 stream, and DeDoDe-G teacher remain
frozen. We construct 14,237 UAV--satellite, satellite--satellite, and
easy-positive pairs with an anchor-disjoint 12,765/1,472
training/validation split. The decoder is distilled for 100 updates
using USAC-MAGSAC-filtered teacher correspondences, symmetric
correspondence InfoNCE, and a preservation loss. Training uses
$448{\times}448$ inputs and inference uses $320{\times}320$ inputs.

Table~\ref{tab:dinov3-rim-control} evaluates the full 3,158-query
Urbanscape set. Every DINOv3 row uses the same Top-5 candidate lists,
2,000 DeDoDe-L keypoints, dual-softmax mutual-nearest descriptor
matching, and a 3-pixel USAC-MAGSAC fundamental-matrix verifier.
``Direct transfer'' means that the previously trained DINOv2 \method\
decoder is executed on DINOv3 tokens without decoder adaptation; it is
not an untrained or randomly initialized local head. This transfer is
already functional because both token maps have 768 channels and the
decoder also receives the common VGG19 multi-scale features. Pair-aware
adaptation increases the mean Top-1 inlier count from 343.35 to 349.11,
while 3D R@1 remains nearly unchanged, with differences of only
+0.063/+0.222 percentage points at 25/50\,m.

\begin{table}[!htbp]
\centering
\caption{Complete DINOv3--\method\ control on all 3,158 Urbanscape
queries. The retrieval checkpoint, Top-5 candidates, detector, matcher,
and verifier are fixed across the DINOv3 rows. Recall@1 (R@1) is measured with the
3D Euclidean protocol. ``DINOv2 decoder, direct transfer'' is a trained
DINOv2--\method\ decoder used without DINOv3-specific adaptation.
Best values are in bold and second-best values are underlined.}
\label{tab:dinov3-rim-control}
\setlength{\tabcolsep}{4pt}
\footnotesize
\begin{tabular}{l|c|cc|c}
\toprule
Variant & Mean Top-1 & \multicolumn{2}{c|}{3D R@1 (\%)} & Mean Top-1 \\
& inliers & 25\,m & 50\,m & distance (m) \\
\midrule
DINOv3 retrieval only & -- & 22.83 & 53.83 & 72.19 \\
DINOv2 decoder, direct transfer & \underline{343.35} & \underline{30.62} & \underline{63.17} & \underline{60.67} \\
\textbf{DINOv3 pair-adapted decoder} & \textbf{349.11} & \textbf{30.68} & \textbf{63.39} & \textbf{60.60} \\
\bottomrule
\end{tabular}
\end{table}

The complete result confirms that retrieval and local matching can share
one DINOv3 token field within \method. Under the current adaptation
protocol, the DINOv3 variant remains below the adopted DINOv2 \method\
result of 34.26/70.80\% at 25/50\,m. After local re-ranking is made
functional, the principal remaining gap is the quality of the cross-domain
DINOv3 retriever. Further adaptation of DINOv3 as the principal \method\
backbone is therefore left for future work.

\FloatBarrier
\section{Controlled Transfer of Sample4Geo Dynamic Similarity Sampling}
\label{app:sample4geo-control}

To compare scenario-dependent training efficiency without changing the
RIM architecture, we transfer Sample4Geo's Dynamic Similarity Sampling
(DSS) loop to the same frozen
DINOv2-B plus SALAD aggregation head. Both alternatives start from the
official SALAD checkpoint, use the reconstructed RIM positive-pair
manifest at $322{\times}322$, freeze the complete DINOv2-B backbone
and global-token projection, and use the same Urbanscape and Park
evaluators. The transferred protocol uses symmetric InfoNCE with
label smoothing 0.1 and a learnable temperature for 40 epochs at
batch size 128. Following the published schedule, geographic neighbours
initialise epochs 1--4; DSS then forms a
Top-128 feature-neighbour pool, selects 64 samples, and refreshes the
embeddings after every four epochs. Within each batch, new seeds and
their selected neighbours are repeatedly added until the batch is full.

\begin{table}[!htbp]
\centering
\caption{Controlled transfer of the Sample4Geo Dynamic Similarity Sampling
(DSS) loop to the frozen Retrieval-In-Matching (RIM)--SALAD head. Park uses the fixed 2,315-query
construction-filtered protocol. Image forwards count training-effective
adaptation and mining/DSS. Retrieval columns are 3D Recall@5 (R@5, \%).}
\label{tab:sample4geo-control}
\setlength{\tabcolsep}{3.4pt}
\scriptsize
\resizebox{\columnwidth}{!}{%
\begin{tabular}{l|rrr|cc|cc}
\toprule
& \multicolumn{3}{c|}{Adaptation compute} & \multicolumn{2}{c|}{Urbanscape} & \multicolumn{2}{c}{Park} \\
Method & Steps & Training forwards & Mining/DSS forwards & 25\,m & 50\,m & 25\,m & 50\,m \\
\midrule
SALAD (no adaptation) & 0 & 0 & 0 & 31.44 & 76.06 & \underline{24.45} & \underline{51.53} \\
RIM two-stage & 1,139 & 72,816 & 49,920 & \textbf{43.26} & \textbf{87.43} & \textbf{31.45} & \textbf{58.96} \\
Sample4Geo DSS-loop control & 880 & 225,280 & 52,164 & \underline{38.19} & \underline{79.16} & 19.83 & 44.28 \\
\bottomrule
\end{tabular}}
\end{table}

The transferred DSS loop uses 277,444 training-effective image forwards,
$2.26\times$ the 122,736 used by the reconstructed RIM two-stage path,
yet trails it by 5.07/8.26\,pp on Urbanscape and 11.62/14.69\,pp on Park
at 25/50\,m.

As a post-hoc training-set diagnostic, we apply RIM's strict criterion to the
DSS neighbour lists; this geographic criterion does not alter DSS
training. At the first feature refresh (after epoch 4), only 22 of 1,360
eligible UAV queries (1.62\%) contain a visually similar but sufficiently
distant candidate. Later training-effective refreshes cover only 3--7
queries (0.22\%--0.51\%). Together with the retrieval results, this
supports a dataset-specific conclusion: repeated full-gallery DSS
refreshes are less cost-effective than one-shot residual-error mining in
our camera-aligned UAV training set. This conclusion is specific to the sparse
hard-negative distribution of this training set; Sample4Geo's original
cross-view datasets have a different negative-sample distribution.

\FloatBarrier
\section{Robustness Boundaries of Geometric Localization}
\label{app:pnp-stress}

Table~\ref{tab:robustness-boundary} tests deterministic query
occlusion on all 3,158 Urbanscape queries with fixed weights. A
centered rectangle covering the specified image-area fraction is
filled with the ImageNet mean RGB value, after which retrieval,
matching, USAC-MAGSAC verification, and PnP are rerun. ``Solver
return'' denotes only that PnP returns a finite pose. Reported medians
are conditional diagnostics over finite solver returns, whereas all
Recall and joint metrics use the full query count as denominator.

\begin{table}[!htbp]
\centering
\caption{Deterministic query-occlusion stress test on Urbanscape.
$e_t$ and $e_R$ denote median 3D translation and rotation errors.}
\label{tab:robustness-boundary}
\setlength{\tabcolsep}{3.0pt}
\scriptsize
\resizebox{\columnwidth}{!}{%
\begin{tabular}{c|cc|c|cc|cc}
\toprule
Occlusion & Retrieval & Re-ranked & Solver & Median $e_t$ & Median $e_R$ & 3\,m/3$^\circ$ & 5\,m/5$^\circ$ \\
(\%) & \multicolumn{2}{c|}{3D Recall@1, 25\,m (\%)} & return (\%) & (m) & ($^\circ$) & (\%) & (\%) \\
\midrule
0  & 26.79 & 33.25 & 100.00 & 2.39 & 1.77 & 56.59 & 79.51 \\
20 & 22.36 & 28.59 & 100.00 & 2.56 & 1.86 & 53.23 & 74.95 \\
40 & 17.07 & 21.06 & 100.00 & 3.19 & 2.18 & 43.89 & 63.52 \\
60 & 11.84 & 14.98 & 100.00 & 5.63 & 3.33 & 30.46 & 45.41 \\
\bottomrule
\end{tabular}}
\end{table}

To isolate the final PnP stage, Table~\ref{tab:pnp-noise} fixes the
unoccluded retrieval result, Top-1 candidate, and cached geometrically
verified correspondences. We then either add fixed-seed zero-mean
Gaussian noise to the query-side 2D coordinates at the original
$720{\times}480$ resolution, or deliberately permute the specified
fraction of query coordinates while retaining their paired 3D points.
PnP is rerun independently for every perturbation level.

\begin{table}[!htbp]
\centering
\caption{Absolute perspective-$n$-point (PnP) failure-boundary sweep with retrieval and matching
fixed. ``Incorrect'' denotes deliberately permuted 2D--3D pairings.
Medians are conditional on a finite solver return; joint metrics use all
queries.}
\label{tab:pnp-noise}
\setlength{\tabcolsep}{2.8pt}
\scriptsize
\resizebox{\columnwidth}{!}{%
\begin{tabular}{ll|c|c|cc|cc}
\toprule
Perturbation & Level & Solver return & Mean inliers & Median $e_t$ & Median $e_R$ & 3\,m/3$^\circ$ & 5\,m/5$^\circ$ \\
& & (\%) & & (m) & ($^\circ$) & (\%) & (\%) \\
\midrule
Gaussian & 0\,px   & 100.00 & 287.52 & 1.42 & 1.21 & 75.93 & 85.43 \\
         & 4\,px   & 100.00 & 219.66 & 2.08 & 1.55 & 65.42 & 82.20 \\
         & 8\,px   & 100.00 & 124.03 & 3.26 & 2.21 & 39.90 & 69.41 \\
         & 16\,px  & 100.00 &  51.43 & 7.50 & 4.57 &  8.39 & 26.98 \\
         & 32\,px  &  99.97 &  20.69 & 17.75 & 10.22 & 0.51 & 2.94 \\
         & 64\,px  &  97.75 &   9.24 & 43.08 & 25.69 & 0.00 & 0.22 \\
         & 128\,px &  70.01 &   3.86 & 75.84 & 51.71 & 0.00 & 0.00 \\
\midrule
Incorrect & 60\%  & 100.00 & 114.90 & 1.60 & 1.31 & 68.24 & 80.49 \\
pairings  & 80\%  &  98.54 &  38.96 & 5.90 & 3.92 & 27.99 & 42.91 \\
          & 90\%  &  88.35 &   7.26 & 86.42 & 60.00 & 1.84 & 3.77 \\
          & 95\%  &  75.55 &   4.59 & 177.82 & 117.07 & 0.09 & 0.22 \\
          & 100\% &  70.08 &   4.10 & 196.59 & 127.21 & 0.00 & 0.00 \\
\bottomrule
\end{tabular}}
\end{table}

\begin{figure}[H]
\centering
\includegraphics[width=0.99\linewidth]{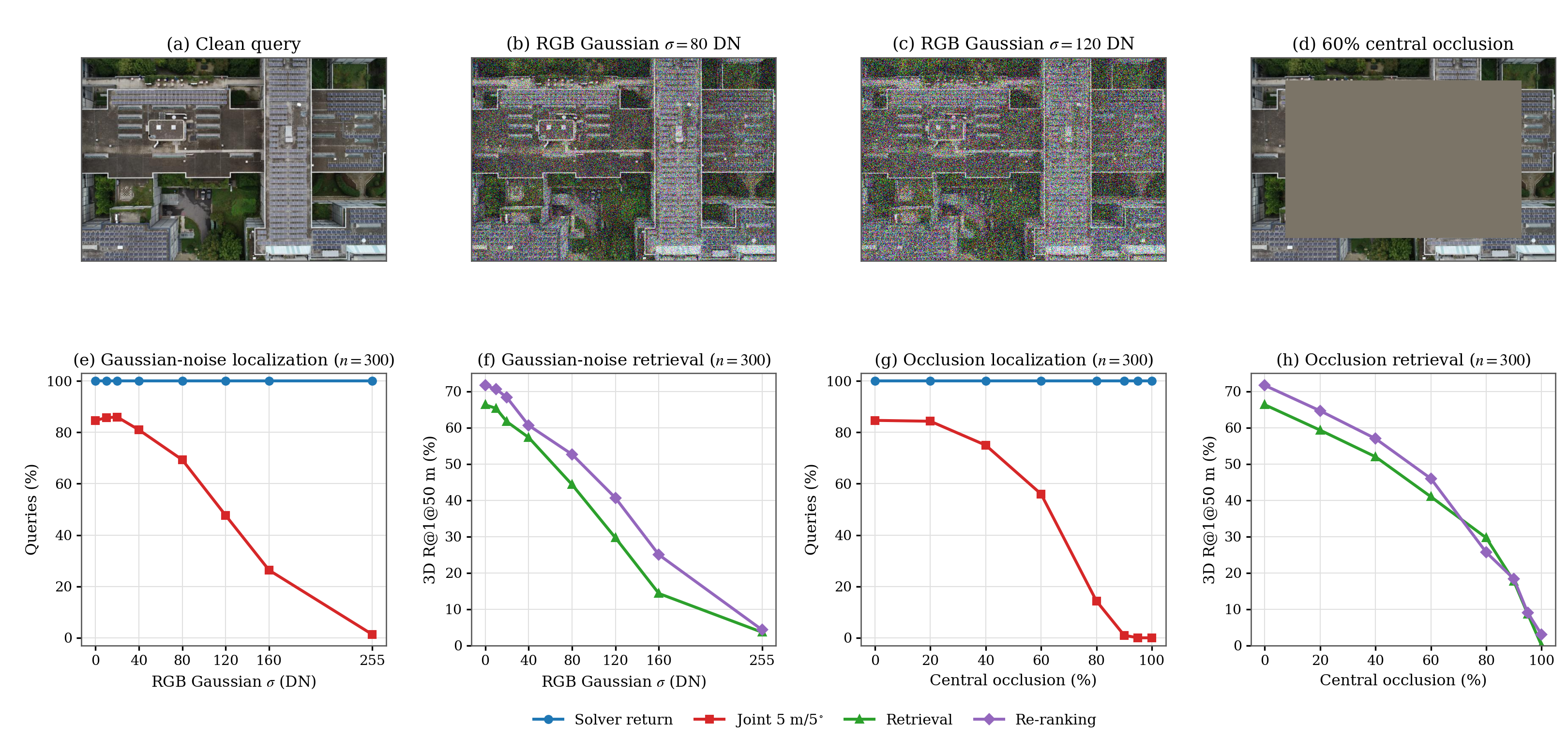}
\caption{Deterministic image-level robustness controls on Urbanscape.
(a)--(d) show a clean query, simulated RGB Gaussian corruption at
$\sigma=80/120$ in 8-bit pixel-value units, and $60\%$ central occlusion. (e)--(f) report the
RGB-noise diagnostic and (g)--(h) the central-occlusion diagnostic on
the same fixed 300-query subset. Each condition reruns full-gallery
retrieval, Top-5 re-ranking, robust verification, and perspective-$n$-point
(PnP) estimation. The retrieval
panels report 3D R@1 at 50\,m.}
\label{fig:pnp-boundary-visual}
\end{figure}

We additionally apply simulated image-level RGB Gaussian corruption before
resizing and rerun the complete pipeline on all 3,158 queries. At
$\sigma=0/80/120$ in 8-bit pixel-value units, solver return remains $100\%$, while joint
$5$\,m/$5^\circ$ accuracy is $85.37/69.38/51.14\%$. Retrieval and
re-ranking 3D R@1 at 50\,m are respectively
$66.34/43.51/29.51\%$ and $69.85/55.45/42.21\%$.
Figure~\ref{fig:pnp-boundary-visual} uses a fixed, evenly spaced
300-query subset to resolve the intermediate degradation trends for both
RGB noise and central occlusion.

The original $100\%$ solver-return rate indicates that \method supplies a
solver-compatible correspondence distribution for every unperturbed
Urbanscape query, but it is distinct from pose correctness. Under $60\%$ central occlusion, joint
$5$\,m/$5^\circ$ accuracy falls to $45.41\%$ and median 3D error reaches
5.63\,m although solver return remains $100\%$; RGB corruption exhibits
the same finite-but-wrong pattern. With retrieval and matching fixed, the
separate estimator control in Table~\ref{tab:pnp-noise} finds repeatable
solver non-return at $\sigma=64$ pixels of artificial query-coordinate
error and at $80\%$ deliberately incorrect 2D--3D pairings. These results
also expose a system limitation: the current pipeline has no calibrated
quality estimate with which to reject a finite but unreliable pose.

\FloatBarrier
\section{Extended Discussion and Park Analysis}
\label{app:extended-discussion}

\paragraph{When does \method underperform the decoupled DeDoDe}
On the re-ranking R@1 columns of Table~\ref{tab:pnp-urbanscape},
the unified \method trails the decoupled DeDoDe teacher by
roughly $1$\,pp. This is the price of sharing a single smaller
(DINOv2-B) backbone between retrieval and matching, rather than
retaining the teacher's dedicated DINOv2-L encoder. Applications
that explicitly require sub-degree / sub-meter precision (e.g.\
inspection landing on a pre-mapped target) may therefore prefer
the decoupled pipeline, whereas \method remains the right
default whenever real-time latency is a first-class concern.

\paragraph{Park error-channel decomposition}
At the tightest joint $5^\circ$/$5$\,m Recall@1 on Chang'an
Park, every pipeline we benchmarked --- including the strongest
dense matchers (RoMa, MatchAnything-RoMa) and the most recent
DINOv3-backed RoMa\,v2 --- remains below $10\%$.
Table~\ref{tab:park-attribution} decomposes this joint number
under the audited datum-corrected ground truth derived from real-time
kinematic (RTK) positioning and exchangeable image file format (EXIF)
metadata into rotation and translation
components. Across all seven methods the rotation-only
R@$5^\circ$ ranges from $50.5\%$ (ALIKED\,+\,LightGlue) to
$71.7\%$ (RoMa) while the position-only R@$5$\,m never exceeds
$7.9\%$. Conditional on the position passing the $5$\,m gate,
the rotation passes the $5^\circ$ gate with $74$--$89\%$
probability for every method, so the joint metric is
arithmetically dominated by R@$5$\,m: the rotation threshold is
largely non-binding, and the observed error is predominantly in the
\emph{translation} channel.

The audited capture geometry is consistent with, but does not by
itself prove, a depth-observability explanation. Urbanscape has a
median/mean off-nadir angle of $30.0^\circ$/$20.66^\circ$, whereas
Park is nearly nadir at $0.10^\circ$/$0.14^\circ$ and 99.74\% of its
queries lie within $5^\circ$. A single near-nadir image provides weak
depth-along-axis constraints, which can contribute to larger 3D
translation error. Temporal appearance change, reference spacing,
mesh quality, and real-time kinematic/camera-metadata uncertainty remain additional possible
contributors; the aggregate table does not isolate their causal
shares.

\begin{table}[!htbp]
\centering
\caption{Park error-channel decomposition under datum-corrected ground
truth derived from real-time kinematic (RTK) positioning and exchangeable
image file format (EXIF) metadata
($N{=}2{,}315$; full test set minus the fixed
68-frame construction-area exclusion,
cf.\ Section~\ref{sec:exp-setup}).
Rotation R@$5^\circ$ is uniformly high while position R@$5$\,m
is the binding constraint of the joint metric. The rightmost
column is the joint $5\,\text{m}/5^\circ$ Recall@1, which
tracks the position channel since R@$5^\circ$ stays high.}
\label{tab:park-attribution}
\setlength{\tabcolsep}{4pt}
\footnotesize
\begin{tabular}{l|cc|cc|cc|c}
\toprule
Method & \shortstack{Median\\3D} & \shortstack{Median\\rotation} & R@$5$\,m & R@$10$\,m & R@$5^\circ$ & R@$10^\circ$ & $5\,\text{m}/5^\circ$ \\
\midrule
RIM (fused-DeDoDe)   & 14.65 & 4.67 &  2.81 & 25.57 & 53.91 & 83.80 &  2.07 \\
DeDoDe (separate)    & 13.81 & 4.46 &  3.07 & 27.43 & 56.72 & 84.62 &  2.46 \\
ALIKED+LightGlue     & 15.47 & 4.96 &  2.89 & 23.56 & 50.48 & 81.70 &  2.50 \\
LoFTR                & 13.62 & 4.45 &  2.90 & 29.26 & 57.99 & 85.57 &  2.59 \\
\textbf{RoMa (v1)}   & \textbf{10.30} & \textbf{3.57} &  7.73 & \textbf{47.08} & \textbf{71.71} & \textbf{90.63} &  6.61 \\
MatchAnything-RoMa   & 10.82 & 3.67 & \textbf{7.86} & 43.07 & 68.34 & 88.86 & \textbf{6.74} \\
RoMa\,v2             & 13.15 & 4.51 &  5.58 & 32.43 & 54.86 & 74.10 &  4.45 \\
\bottomrule
\end{tabular}
\end{table}

\paragraph{Simulated altitude-prior diagnostic}
To test the sensitivity of the translation channel to independent
altitude information, we simulate an idealised altitude prior of
varying precision $\sigma_b$. For each query we draw
$\hat{h}_b \sim \mathcal{N}(h_\text{GT}, \sigma_b^2)$ along
the local-up direction and Bayesian-fuse it with the
single-view PnP altitude under a fixed sensitivity setting
$\sigma_z = 8$\,m, then update the camera position by the
posterior shift along $\hat{u}$
(\ref{app:park-altprior}). The fusion is post-hoc and we average
over five fixed random seeds. No physical barometer measurements
are used.

\begin{table}[!htbp]
\centering
\caption{Joint $5\,\text{m}/5^\circ$ Recall@1 (R@1, \%) on Park as a
function of the simulated independent altitude-noise level
$\sigma_b$ (m). The sweep is a sensitivity analysis, not an
evaluation of a particular barometer or fusion implementation.}
\label{tab:park-altprior}
\setlength{\tabcolsep}{4pt}
\footnotesize
\begin{tabular}{l|c|cccccc}
\toprule
Method & no prior & $\sigma_b\!=\!1$ & $\sigma_b\!=\!2$ & $\sigma_b\!=\!5$ & $\sigma_b\!=\!10$ & $\sigma_b\!=\!20$ & $\sigma_b\!=\!50$ \\
\midrule
ALIKED+LightGlue     &  2.50 & 11.22 & 10.24 &  7.02 &  5.56 &  4.05 &  2.88 \\
DeDoDe (separate)    &  2.46 & 14.88 & 13.28 &  8.66 &  6.70 &  4.79 &  3.01 \\
LoFTR                &  2.59 & 14.82 & 13.30 &  9.14 &  6.78 &  4.68 &  3.25 \\
RIM (fused-DeDoDe)   &  2.07 & 11.36 & 10.49 &  6.79 &  5.09 &  4.06 &  2.80 \\
\textbf{RoMa (v1)}   &  6.61 & \textbf{30.51} & \textbf{28.10} & \textbf{18.92} & \textbf{14.42} & \textbf{10.32} &  7.37 \\
MatchAnything-RoMa   & \textbf{6.74} & 27.31 & 25.27 & 17.61 & 13.39 &  9.83 & \textbf{7.45} \\
RoMa\,v2             &  4.45 & 19.29 & 17.76 & 11.91 &  8.95 &  6.78 &  4.81 \\
\midrule
relative gain ($\sigma_b{=}1$) & --- & $4.0\!-\!6.1\!\times$ & & & & & $\sim\!1\!\times$\\
\bottomrule
\end{tabular}
\end{table}

The monotonic gains show that the Park metric is sensitive to an
independent altitude cue, especially at the strict 5\,m position
threshold. This supports investigating real sensor fusion, but the
simulation does not establish the performance of a physical
barometer, prove a unique geometric cause, or replace the reported
visual-only operating point.

\FloatBarrier
\section{Re-ranking depth $K$ scaling and reduced-precision results}
\label{app:latency-scaling}

This appendix gives the full per-query computational-inference
latency as a function of the re-ranking depth $K$, at both
32-bit floating point (FP32, the main-text regime of
Table~\ref{tab:latency}) and 16-bit floating point (FP16)
on the query-side forwards. All numbers are measured on the
Urbanscape test split (averaged over $100$ deterministically
sampled queries), on a single NVIDIA GeForce RTX 3090 (24\,GB),
at input resolution $322{\times}322$, processing \emph{one
query at a time} (no query-side or matcher-side batching),
with device-side stages separated by explicit CUDA
synchronisation so that every reported on-device time
reflects finished-kernel latency rather than command-queue
dispatch. This single-query protocol matches the on-board-UAV
deployment assumption of Section~\ref{sec:exp-latency} and is
therefore not directly comparable to batched throughputs
reported in prior work. All sparse matchers use USAC-MAGSAC
for geometric verification (identical to the accuracy
evaluator). Image input/output, resizing, host-to-device transfer, and the database-side
encoder are excluded; the FP32 columns at $K{=}5$ reproduce
the \emph{Total} column of Table~\ref{tab:latency} exactly.

\begin{table*}[!htbp]
\centering
\caption{Per-query latency (ms) versus re-ranking depth $K$,
at 32-bit floating point (FP32) and 16-bit floating point (FP16). Total $= t_{\text{setup,gpu}} +
K\cdot(t_{\text{match,gpu}}^{\text{pair}}
+ t_{\text{match,cpu}}^{\text{pair}})$, matching the
\emph{Total} column of Table~\ref{tab:latency} at $K{=}5$.
$\partial T/\partial K$ is the per-candidate total matcher
cost. Same protocol as Table~\ref{tab:latency}.
MNN denotes mutual nearest-neighbor matching.
EfficientLoFTR FP16 was not included in the native-implementation
audit. RoMa\,FP16 is marked not available (N/A) because its internal linear solver
fails on a \texttt{Half}/\texttt{Float} dtype mismatch under
\texttt{torch.autocast}, so no pair yields valid
correspondences.}
\label{tab:latency-scaling}
\setlength{\tabcolsep}{4pt}
\footnotesize
\resizebox{\linewidth}{!}{%
\begin{tabular}{l|cccc|cccc}
\toprule
& \multicolumn{4}{c|}{\textbf{FP32}}
& \multicolumn{4}{c}{\textbf{FP16} (query-side autocast)} \\
Pipeline
   & $K{=}5$ & $K{=}10$ & $K{=}20$ & $\partial T/\partial K$
   & $K{=}5$ & $K{=}10$ & $K{=}20$ & $\partial T/\partial K$ \\
\midrule
Ours$^{\ast}$\,+\,ALIKED\,+\,LightGlue & 160.9 &  289.9 &  548.0 &  25.81 & 174.3 &  322.7 &  619.6 &  29.69 \\
Ours$^{\ast}$\,+\,ALIKED\,+\,MNN       &  38.0 &   44.2 &   56.5 &   1.23 &  33.2 &   39.8 &   52.9 &   1.31 \\
Ours$^{\ast}$\,+\,LoFTR                 & 410.7 &  799.9 & 1578.4 &  77.85 & 268.2 &  522.4 & 1030.7 &  50.83 \\
Ours$^{\ast}$\,+\,EfficientLoFTR        & 164.2 &  307.0 &  592.6 &  28.56 & \multicolumn{4}{c}{\emph{N/A (not measured)}} \\
Ours$^{\ast}$\,+\,DeDoDe (separate)    & 111.7 &  169.6 &  285.3 &  11.57 & 121.7 &  182.8 &  304.9 &  12.21 \\
Ours$^{\ast}$\,+\,RoMa                 & 2746.7 & 5474.9 & 10931.3 & 545.64 & \multicolumn{4}{c}{\emph{N/A (see caption)}} \\
\midrule
\rowcolor{gray!12}
\textbf{Ours$^{\ast}$\,+\,\method (unified)}
                                       &  90.8 &  143.4 &  248.6 &   10.52 &  61.5 &   96.8 &  167.3 &   7.05 \\
\bottomrule
\end{tabular}}
\end{table*}

\paragraph{How to read this table}
The FP32 half of the table decomposes the cost-model
predictions of
Eqs.~\eqref{eq:rim-cost-baselines}--\eqref{eq:rim-cost-ours}.
The decoupled DeDoDe pipeline's total latency grows as
$t_{\text{global}} + t_{\text{det}}^{q} + t_{\text{desc}}^{q}
+ K\cdot t_{\text{MNN}}^{\text{pair}}$, whereas \method grows
as $t_{\text{shared}} + t_{\text{det}}^{q} +
t_{\text{head}} + K\cdot t_{\text{MNN}}^{\text{pair}}$
($t_{\text{shared}} = t_{\text{DINOv2\text{-}B}} + t_{\text{VGG19}}$):
\method trades the DeDoDe-separate global encoder
$t_{\text{global}}$ for the (smaller) combined
DINOv2-B\,+\,head forward, and remains in the same shallow-$K$
regime as DeDoDe-separate because both pipelines run
Dual-SoftMax matching per candidate. Pairwise matchers, by contrast, have
$T(K) = t_{\text{global}} + K\cdot t_{\text{match}}^{\text{pair}}$
with $t_{\text{match}}^{\text{pair}}$ being their full
per-pair forward, which gives the much steeper slopes
observed in the table (EfficientLoFTR $28.6$, LoFTR $77.8$,
RoMa $545.6$\,ms per candidate). The resulting FP32 ordering is
ALIKED\,+\,MNN $<$ \method $<$ DeDoDe-separate $<$
ALIKED\,+\,LightGlue $<$ EfficientLoFTR $<$ LoFTR $\ll$ RoMa
at every $K$. The ablation
Ours$^{\ast}$\,+\,ALIKED\,+\,MNN is the only row faster than
\method on Total, but recovers a valid PnP solution on only
$71.4\%$ of queries (Table~\ref{tab:pnp-urbanscape}) and thus
operates at a qualitatively different point on the
latency\,/\,quality Pareto frontier.

\paragraph{Reduced-precision observations}
Switching query-side forwards to automatic mixed-precision
(float16) gives the expected speed-up for the dense matchers
at saturation (LoFTR $1.5{\times}$ faster on Total) and for
the ALIKED\,+\,MNN lightweight baseline ($1.2{\times}$), while
DeDoDe-separate changes negligibly because its CPU-USAC share
is already large. \method also benefits moderately ($1.1{\times}$
at $K{=}5$). ALIKED\,+\,LightGlue is slightly \emph{slower} in
FP16, a pattern we attribute to the half-/single-precision dtype
casts that the autocast context inserts between compute-bound
and small non-autocast-listed kernels; the overhead outweighs
the arithmetic savings for this pipeline. RoMa is
reported as N/A because every probe-pair forward raises a
dtype mismatch inside its internal linear solver (see
caption); consequently, RoMa is excluded from the FP16
column rather than reporting a truncated number. Across all
remaining pipelines the \emph{ordering} of pipelines is
identical in FP32 and FP16, so the main-text conclusions drawn
from Table~\ref{tab:latency} carry over unchanged to the FP16
regime.

\FloatBarrier
\section{Planar (XY) retrieval tables}
\label{app:xy-tables}

This appendix reports the planar (XY) distance counterparts of
the main 3D retrieval tables
(Tables~\ref{tab:rec-urbanscape-3d} and~\ref{tab:rec-park-3d}).
We moved the XY variants here because they tell essentially the
same story as the 3D metric --- same method ordering, same
interaction pattern of fine-tuned vs.\ off-the-shelf baselines,
only with uniformly higher absolute numbers because altitude
errors are not penalised. For completeness,
Tables~\ref{tab:rec-urbanscape-xy} and
\ref{tab:rec-park-xy} give the full XY results at
$\tau\in\{25,50\}$\,m under the same conventions as the 3D
tables in the main text.

% --- Table A1: Urbanscape XY ---
\begin{table*}[!htbp]
\centering
\caption{Recall@$K$ (\%) on the EPFL urban (\textit{Urbanscape})
benchmark, evaluated with the planar (XY) distance metric at
tolerance radii $\tau\in\{25,50\}$\,m. Conventions follow the
main 3D version in Table~\ref{tab:rec-urbanscape-3d}.
``$\ast$'' marks models whose retrieval head is retrained on
our pose-near UAV-reference pairs.}
\label{tab:rec-urbanscape-xy}
\setlength{\tabcolsep}{4pt}
\scriptsize
\resizebox{\linewidth}{!}{%
\begin{tabular}{l|c|cc|cc}
\toprule
& & \multicolumn{2}{c|}{$\tau = 25$\,m} & \multicolumn{2}{c}{$\tau = 50$\,m} \\
Method & Dim. & R@1 & R@5 & R@1 & R@5 \\
\midrule
HLoc (AP-GeM)~\citep{sarlin2019hloc,revaud2019apgem} & $2048$   & 17.42 & 60.04 & 66.72 & 85.40 \\
CLIP~\citep{radford2021clip}                          & $768$    & 6.62  & 20.87 & 21.31 & 47.63 \\
CosPlace~\citep{berton2022cosplace}                   & $2048$   & 26.12 & 49.46 & 52.98 & 76.41 \\
DINOv2 \texttt{[CLS]}~\citep{oquab2024dinov2}         & $768$    & 18.14 & 40.69 & 45.03 & 73.15 \\
EigenPlaces~\citep{berton2023eigenplaces}             & $2048$   & 35.28 & 63.33 & 70.08 & 84.86 \\
AnyLoc~\citep{keetha2023anyloc}                       & $49152$  & 36.35 & 65.90 & 70.80 & 91.74 \\
SALAD~\citep{izquierdo2024salad}                      & $8448$   & 36.64 & 65.29 & \underline{76.79} & \underline{92.69} \\
SelaVPR++~\citep{lu2025selavprpp}                     & $2048$   & 35.24 & 65.90 & 73.97 & 91.17 \\
Game4Loc (zero-shot)~\citep{ji2025game4loc}               & $768$    & 26.63 & 52.15 & 54.65 & 78.53 \\
\midrule
NetVLAD (DINOv2)$^{\ast}$~\citep{arandjelovic2016netvlad} & $49152$  & 22.29 & 45.47 & 54.27 & 77.11 \\
SelaVPR++$^{\ast}$~\citep{lu2025selavprpp}                & $2048$   & \underline{36.38} & \underline{67.48} & 71.63 & 88.41 \\
\rowcolor{gray!12}
\textbf{Ours}$^{\ast}$                                    & $8448$
  & \textbf{44.87}\,\scriptsize{\textcolor{red}{$+8.23$}}
  & \textbf{74.16}\,\scriptsize{\textcolor{red}{$+8.87$}}
  & \textbf{85.31}\,\scriptsize{\textcolor{red}{$+8.52$}}
  & \textbf{94.87}\,\scriptsize{\textcolor{red}{$+2.18$}} \\
\bottomrule
\end{tabular}}\\[2pt]
\parbox{0.95\linewidth}{\centering\scriptsize Red deltas: gain
of \textbf{Ours}$^{\ast}$ over the vanilla SALAD backbone.}
\end{table*}

% --- Table A2: Park XY ---
\begin{table*}[!htbp]
\centering
\caption{Recall@$K$ (\%) on the Chang'an Park benchmark, planar
(XY) distance metric, on the $2{,}315$-query
construction-filtered set (matching
Table~\ref{tab:park-attribution}). Conventions follow
Table~\ref{tab:rec-urbanscape-xy}.}
\label{tab:rec-park-xy}
\setlength{\tabcolsep}{4pt}
\scriptsize
\resizebox{\linewidth}{!}{%
\begin{tabular}{l|c|cc|cc}
\toprule
& & \multicolumn{2}{c|}{$\tau = 25$\,m} & \multicolumn{2}{c}{$\tau = 50$\,m} \\
Method & Dim. & R@1 & R@5 & R@1 & R@5 \\
\midrule
HLoc (AP-GeM)~\citep{sarlin2019hloc,revaud2019apgem} & $2048$  & 17.93 & 39.83 & 30.97 & 62.46 \\
CLIP~\citep{radford2021clip}                          & $768$   & 6.26  & 21.38 & 18.27 & 48.21 \\
CosPlace~\citep{berton2022cosplace}                   & $2048$  & 9.72  & 26.48 & 18.57 & 46.09 \\
DINOv2 \texttt{[CLS]}~\citep{oquab2024dinov2}         & $768$   & 14.25 & 33.56 & 31.32 & 62.81 \\
EigenPlaces~\citep{berton2023eigenplaces}             & $2048$  & 17.49 & 34.90 & 28.34 & 56.37 \\
AnyLoc~\citep{keetha2023anyloc}                       & $49152$ & 17.49 & 44.88 & 34.17 & 74.34 \\
SALAD~\citep{izquierdo2024salad}                      & $8448$  & 17.37 & 40.30 & 31.23 & 67.95 \\
SelaVPR++~\citep{lu2025selavprpp}                     & $2048$  & \underline{21.99} & \underline{46.05} & \textbf{42.76} & \underline{75.16} \\
Game4Loc (zero-shot)~\citep{ji2025game4loc}               & $768$   & 16.89 & 40.26 & 31.06 & 68.47 \\
\midrule
NetVLAD (DINOv2)$^{\ast}$~\citep{arandjelovic2016netvlad} & $49152$ & 14.69 & 32.27 & 30.97 & 55.64 \\
SelaVPR++$^{\ast}$~\citep{lu2025selavprpp}                & $2048$  & 18.53 & 42.25 & 41.12 & 72.35 \\
\rowcolor{gray!12}
\textbf{Ours}$^{\ast}$                                    & $8448$
  & \textbf{23.28}\,\scriptsize{\textcolor{red}{$+5.91$}}
  & \textbf{48.55}\,\scriptsize{\textcolor{red}{$+8.25$}}
  & \underline{41.38}\,\scriptsize{\textcolor{red}{$+10.15$}}
  & \textbf{77.80}\,\scriptsize{\textcolor{red}{$+9.85$}} \\
\bottomrule
\end{tabular}}\\[2pt]
\parbox{0.95\linewidth}{\centering\scriptsize Red deltas: gain
of \textbf{Ours}$^{\ast}$ over the vanilla SALAD backbone.}
\end{table*}

\FloatBarrier
\section{Cross-domain evidence from capture statistics}
\label{app:capture-stats}

The cross-domain nature of the two benchmarks introduced in
Section~\ref{sec:exp-setup} is not only qualitative. The two
benchmarks differ in how the temporal gap between reference
and query is induced. On Urbanscape, the photogrammetric mesh
served by Google for any given area is typically reconstructed
from imagery spanning several years, whereas the real UAV
queries are acquired in a bounded multi-season window. On
Chang'an Park, both sides are flown by the authors with DJI
multirotors, but the SfM reference flight predates the
evaluation flight by at least one year, producing a comparable
year-plus offset of explicitly known provenance. In either
case the resulting temporal offset is a primary driver of the
appearance shift addressed by Stage~1 adaptation.
Table~\ref{tab:capture-stats} quantifies this offset for the
two benchmarks: Urbanscape queries are acquired over a
multi-season window (three flight dates in Autumn 2020) while
Chang'an Park queries come from two consecutive-day flights
in Autumn 2023 (\texttt{2023-10-20} 11:19:41 through
\texttt{2023-10-21} 17:11:14, China local time, recovered
from the per-frame DJI exchangeable image file format (EXIF)
\texttt{CreateDate} on the
unprocessed raw imagery); in both cases the gap between
query capture time and reference mesh age is large enough
that query-vs-reference appearance drift is material rather
than cosmetic.

\begin{table}[!htbp]
\centering
\caption{Query capture-time statistics of the two benchmarks.
Reference imagery is served by Google Photorealistic 3D
Tiles~\citep{google3dtiles2024} for Urbanscape and by a
self-reconstructed structure-from-motion (SfM) mesh (built from a separate DJI oblique
flight) for Chang'an Park; in both cases the gap between query
capture and reference vintage materially exceeds one year.}
\label{tab:capture-stats}
\setlength{\tabcolsep}{6pt}
\footnotesize
\begin{tabular}{l|cc}
\toprule
Benchmark & Urbanscape & Chang'an Park \\
\midrule
\# real UAV queries ($N_q$)             & $3{,}158$          & $2{,}383$ \\
Reference mesh source                   & Google 3D Tiles & Self SfM (DJI) \\
Capture year(s) (queries)               & $2020$             & $2023$    \\
Capture span                            & $\approx\!2$\,months & $\approx\!2$\,days \\
Reference-vs-query temporal gap (years) & $\geq\!3$          & $\geq\!1$ \\
\bottomrule
\end{tabular}
\end{table}

\FloatBarrier
\section{Qualitative failure cases}
\label{app:failure-cases}

\begin{figure*}[!htbp]
\centering
\includegraphics[width=0.82\linewidth]{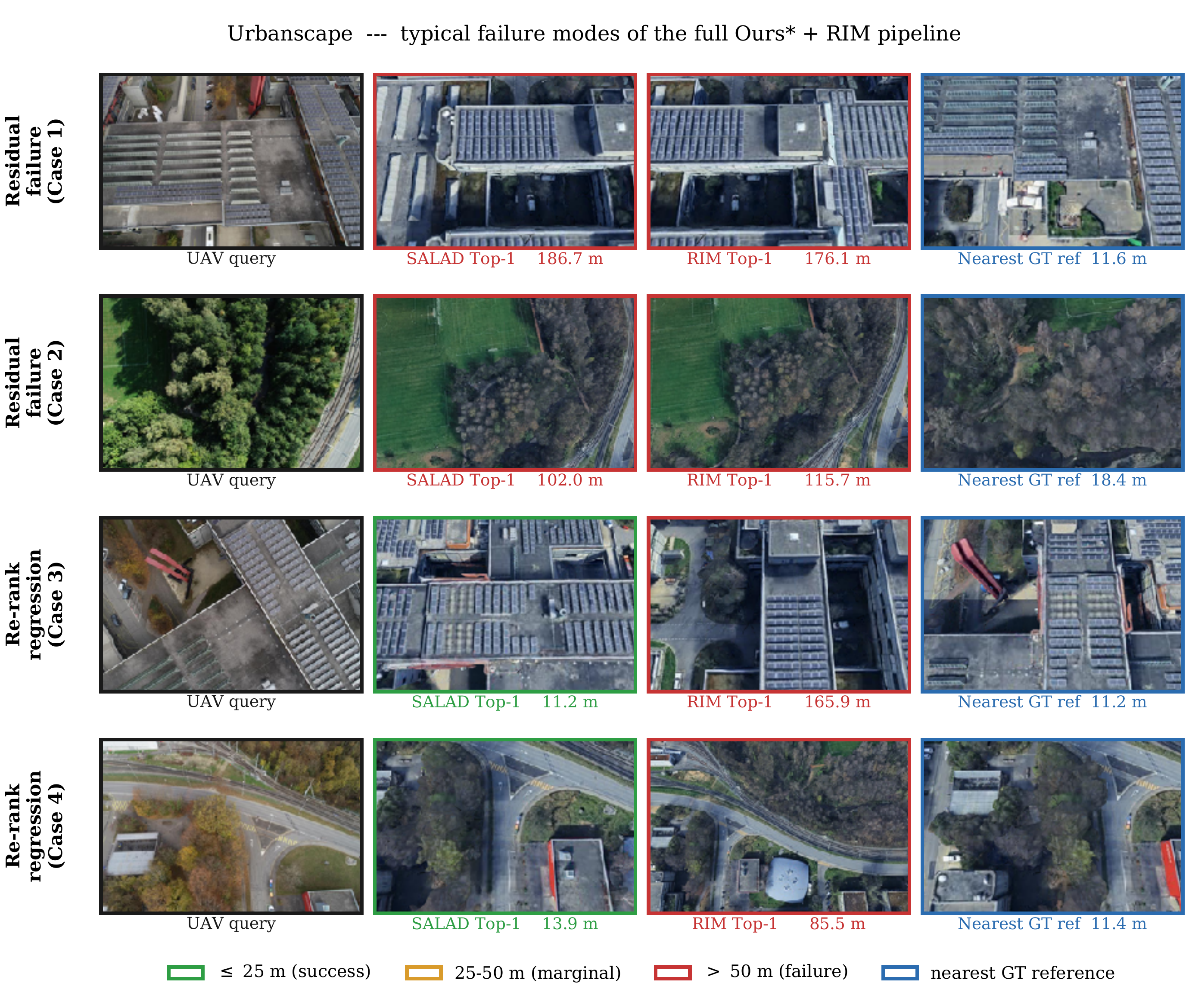}
\caption{Typical failure cases of the full
Ours$^{\ast}$\,+\,\method pipeline on Urbanscape. Each row
shows a UAV query (leftmost) followed by the retriever Top-$K$
(green/red borders mark references within/outside the $25$\,m
tolerance). Failures cluster around architectural
self-similarity, strong query-vs-reference appearance drift,
and boundary queries with thin reference support; a detailed
analysis is given in \ref{app:failure-cases}.}
\label{fig:failure-cases}
\end{figure*}

To keep the main text focused on the quantitative story we
collect a handful of failure cases in this appendix.
Figure~\ref{fig:failure-cases} shows Top-1 retrievals produced
by the full Ours$^{\ast}$\,+\,\method pipeline on Urbanscape
queries for which the XY pose error exceeds $50$\,m, together
with the ground-truth reference tile. Three systematic
patterns emerge. \emph{(i) Self-similar architectural
repetition}: nearby blocks of rooftops with near-identical
shape and orientation can produce a descriptor whose
cosine similarity with the incorrect tile is higher than with
the correct one; retrieval succeeds in recovering the
``correct type'' of tile but not the correct instance.
\emph{(ii) Strong appearance drift at a visually distinctive
landmark}: a handful of queries that front a construction
site or seasonal vegetation receive reference-side imagery
from a prior year in which the same region looks
qualitatively different. \emph{(iii) Bounded support at the
benchmark edge}: queries captured near the edge of the 3D
tile footprint have a thin shell of valid references, so a
single catastrophic retrieval error cannot be overridden by
an equally strong second-best. In all three regimes the
downstream re-ranking and PnP stages cannot recover the
correct pose, which motivates future work on
retrieval-side hardening (e.g.\ multi-scale retrieval, or
graph-level aggregation of overlapping reference tiles).

\FloatBarrier
\section{Park evaluation-set audit}
\label{app:park-valid-regions}

The Park flight contains 2,383 query frames. We apply the fixed
construction-area exclusion described in Section~\ref{sec:exp-setup}
and evaluate the remaining 2,315 queries. The exclusion is necessary
because those 68 views depict newly constructed content that has no
counterpart in the year-older reference mesh; it is therefore an
open-set map-coverage filter, not a statement that RIM can localise
against obsolete content. The full flight spans 131.21\,m in World Geodetic System 1984
ellipsoidal height and 20.86\,km after summing consecutive sub-25\,m
pose steps. Moreover, 99.74\% of the retained cameras lie within
$5^\circ$ of nadir (median $0.10^\circ$, mean $0.14^\circ$). These
audited properties motivate treating Park as a near-nadir,
cross-temporal stress test, without assigning an unsupported numeric
ground-truth precision floor.

\FloatBarrier
\section{Park Datum-Corrected Ground Truth}
\label{app:park-gt}

The Park evaluation uses translations derived from dual-frequency
real-time kinematic (RTK) positioning and rotations parsed from the
camera exchangeable image file format (EXIF) metadata.
To align the query-flight and reference-mesh datums, the released
evaluation pose array applies the same ECEF translation
$[-2.4223,\,7.0706,\,5.0688]$\,m to every query. The standard
deviation of this correction across queries is zero in all three
axes, and every rotation matrix is bit-identical to the original
EXIF-derived pose array. Consequently, the evaluated protocol is a
constant datum correction of RTK/EXIF ground truth; it does not mix
matcher-derived ``Oracle'' rotations with EXIF rotations. This direct
array comparison is included in the released protocol-audit script so
that the transformation and query count can be reproduced exactly.

The corrected poses make the query and mesh coordinate frames
consistent, but they do not remove normal sensor uncertainty or the
weak depth observability of a single near-nadir view. We therefore
report both translation and rotation errors and avoid interpreting
the correction itself as a learned or image-derived improvement.

\FloatBarrier
\section{Park altitude-prior fusion: diagnostic protocol}
\label{app:park-altprior}

The supplemental altitude-prior ablation is a simulation rather than
a measurement from an onboard barometer. For every test query $q$ we
keep the cached PnP $2$D--$3$D correspondences and refined camera pose
$\mathbf{T}_q^\text{PnP}\in SE(3)$; only its position is adjusted.
Let $\hat{\mathbf{u}}$ denote the local-up direction at the dataset
ECEF origin and let
$h^\text{PnP}_q=\langle\mathbf{t}_q^\text{PnP}-\mathbf{p}_0,
\hat{\mathbf{u}}\rangle$. We use $\sigma_z=8$\,m as a fixed
sensitivity setting for PnP altitude uncertainty and draw
\begin{equation}
\hat{h}^b_q\sim\mathcal{N}\!\left(h^\text{GT}_q,\sigma_b^2\right),
\end{equation}
where $h^\text{GT}_q$ is the datum-corrected RTK evaluation altitude.
The two independent Gaussian readings are fused as
\begin{equation}
h^\text{post}_q=\frac{\sigma_b^2 h^\text{PnP}_q+
\sigma_z^2\hat{h}^b_q}{\sigma_b^2+\sigma_z^2},\qquad
\Delta h_q=h^\text{post}_q-h^\text{PnP}_q,
\end{equation}
and $\mathbf{t}_q^\text{post}=\mathbf{t}_q^\text{PnP}+
\Delta h_q\hat{\mathbf{u}}$; rotation is unchanged. We average five
fixed Monte-Carlo seeds for each $\sigma_b$. The result quantifies
sensitivity to an idealised independent altitude cue. It is not a
claim about a particular barometer, nor a lower bound on a future
tightly coupled estimator.

\FloatBarrier
\section{Cross-backbone specificity of the orthogonal recipe}
\label{app:cross-backbone}

A natural question is whether the gains of Ours$^{\ast}$ simply
encode ``train on the right database with the right loss'', in
which case analogous adaptation on the same data should help
another foundation-model retriever equally well. We therefore apply
the same UAV--reference pairs and two-stage easy-positive/hard-negative
structure to
SelaVPR++~\citep{lu2025selavprpp}, a strong recent
DINOv2-based retriever whose aggregation is built on a
\texttt{[CLS]}-token plus local-reranking stream rather than on
SALAD's optimal-transport soft-assignment. The architecture-specific
optimisation schedules are not identical: the recorded SelaVPR++
Stage~1 run uses two epochs, whereas SALAD uses one, and the freeze
policies necessarily target different modules. Table~\ref{tab:recipe-specificity}
is therefore a same-data exposure study, not a single-factor
cross-backbone control.

\begin{table*}[!htbp]
\centering
\caption{Same-data exposure study of two-stage adaptation on
SelaVPR++ and SALAD. Both use the same UAV--reference pairs and
easy-positive/hard-negative structure, but retain architecture-specific
schedules and freeze policies. The comparison therefore does not
isolate pretraining, optimisation, or aggregation architecture as a
single causal factor. Numbers are Recall@$1$ (\%); deltas are the
finetuned-vs-original gap for each backbone (green: gain;
red: degradation). For each benchmark we report XY and 3D
distance variants at $\tau\in\{25,50\}$\,m; Park numbers are
on the $2{,}315$-query construction-filtered set (matching
Table~\ref{tab:rec-park-3d}).}
\label{tab:recipe-specificity}
\setlength{\tabcolsep}{3pt}
\scriptsize
\resizebox{\linewidth}{!}{%
\begin{tabular}{l|cc|cc||cc|cc}
\toprule
& \multicolumn{4}{c||}{Urbanscape R@1 (\%)}
& \multicolumn{4}{c}{Chang'an Park R@1 (\%)} \\
& \multicolumn{2}{c|}{XY}
& \multicolumn{2}{c||}{3D}
& \multicolumn{2}{c|}{XY}
& \multicolumn{2}{c}{3D} \\
Variant
& $\tau{=}25$ & $\tau{=}50$ & $\tau{=}25$ & $\tau{=}50$
& $\tau{=}25$ & $\tau{=}50$ & $\tau{=}25$ & $\tau{=}50$ \\
\midrule
SelaVPR++ (Original)
  & 35.24 & 73.97 & 18.52 & 54.84
  & 21.99 & 42.76 & 12.10 & 30.89 \\
SelaVPR++ (adapted$^{\ast}$)
  & 36.38 & 71.63 & 22.17 & 58.49
  & 18.53 & 41.12 &  9.24 & 27.78 \\
\addlinespace[1pt]
$\;\;\Delta$ (recipe $\rightarrow$ SelaVPR++)
  & \textcolor{OliveGreen}{$+1.14$}
  & \textcolor{red}{$-2.34$}
  & \textcolor{OliveGreen}{$+3.65$}
  & \textcolor{OliveGreen}{$+3.65$}
  & \textcolor{red}{$-3.46$}
  & \textcolor{red}{$-1.64$}
  & \textcolor{red}{$-2.86$}
  & \textcolor{red}{$-3.11$} \\
\midrule
SALAD (Original)
  & 36.64 & 76.79 & 18.43 & 53.93
  & 17.37 & 31.23 & 10.63 & 23.24 \\
\rowcolor{gray!12}
\textbf{SALAD (Ours$^{\ast}$)}
  & \textbf{44.87} & \textbf{85.31} & \textbf{26.98} & \textbf{67.70}
  & \textbf{23.28} & \textbf{41.38} & \textbf{15.08} & \textbf{32.18} \\
\addlinespace[1pt]
$\;\;\Delta$ (recipe $\rightarrow$ SALAD)
  & \textcolor{OliveGreen}{$+8.23$}
  & \textcolor{OliveGreen}{$+8.52$}
  & \textcolor{OliveGreen}{$+8.55$}
  & \textcolor{OliveGreen}{$+13.77$}
  & \textcolor{OliveGreen}{$+5.91$}
  & \textcolor{OliveGreen}{$+10.15$}
  & \textcolor{OliveGreen}{$+4.45$}
  & \textcolor{OliveGreen}{$+8.94$} \\
\bottomrule
\end{tabular}}
\end{table*}

The empirical pattern is consistent on \emph{both} benchmarks.
The two-stage SALAD adaptation lifts SALAD by $+4$
to $+13.8$\,pp across all eight R@1 columns, while analogous adaptation
at most produces marginal, sign-indefinite changes for
SelaVPR++ on Urbanscape ($-2.3$ to $+3.7$\,pp) and actively
degrades it across every Park column ($-1.6$ to $-3.5$\,pp).
Because both backbones consume the same UAV--remote-sensing pairs,
different dataset exposure does not explain the gap. The experiment
does not, however, separately identify the effects of source
pretraining, aggregation architecture, schedule, freeze policy, or
hyper-parameter compatibility. A
plausible interpretation, anticipated in
Section~\ref{sec:method-finetune}, is that the two stages of our
recipe are well matched to SALAD's
cluster\,+\,global-token factorisation: Stage~1 reuses SALAD's
optimal-transport aggregation as an easy-positive similarity
anchor, while Stage~2 refines the cluster branch on a small
set of mined hard negatives without touching the frozen
scene-level global token. SelaVPR++'s aggregation does not share
this decomposition, which is consistent with its weaker transfer
under this recipe. Establishing whether its Park degradation is
caused by aggregation, pretraining-domain mismatch or recipe-
specific overfitting requires a controlled component study and is
therefore left as a hypothesis rather than a causal conclusion.

\FloatBarrier

\end{document}